%% file: submission.tex
\documentclass{article}  

\input{settings/preamble.tex}

\begin{document}
\input{settings/author-info.tex}
\input{sections/abstract/main.tex}

\input{sections/introduction/main.tex}
\input{sections/background/main.tex}

\input{sections/methods/main.tex}

\input{sections/experiments/main.tex}

\input{sections/related-work/main.tex}
\input{sections/conclusions/main.tex}

\clearpage
\input{sections/misc/acknowledgements.tex}

\bibliographystyle{bib-style}  
\bibliography{ref}

\input{settings/appendix-information.tex}

\end{document}

%% file: settings/preamble.tex
\input{settings/packages.tex}
\input{settings/layout.tex}  
\input{settings/paper-status.tex}

\input{settings/user-commands.tex}

%% file: settings/packages.tex
\usepackage{ijcai22}

\usepackage{times}
\usepackage{soul}
\usepackage[small]{caption}
\usepackage{graphicx}
\usepackage{booktabs}

\usepackage[hidelinks]{hyperref}  
\usepackage[utf8]{inputenc}
\usepackage{url}

\usepackage{amsmath}
\usepackage{amssymb}
\usepackage{bbm}
\usepackage{bm}
\usepackage{cancel}
\usepackage{mathtools}

\usepackage{ascmac}  
\usepackage{algorithm}
\usepackage{algpseudocode}

\usepackage{fancyhdr}
\usepackage{tcolorbox}
\usepackage{tikz}

\usepackage{booktabs}
\usepackage{comment}
\usepackage{lastpage}
\usepackage{listings}
\usepackage{multibib}
\usepackage{multirow}
\usepackage[super]{nth}
\usepackage[caption=false]{subfig}

\allowdisplaybreaks

\tcbuselibrary{breakable, skins, theorems}

\newtheorem{theorem}{Theorem}

\newtheorem{corollary}{Corollary}

\newtheorem{proposition}{Proposition}

\newtheorem{lemma}{Lemma}

\newtheorem{definition}{Definition}

\newtheorem{proof}{Proof}

\algtext*{EndFor}
\algtext*{EndWhile}
\algtext*{EndIf}
\algtext*{EndProcedure}
\algtext*{EndFunction}
\algnewcommand{\LineComment}[1]{\State \(\triangleright\) #1}

\newcites{appx}{References}

%% file: settings/layout.tex
\fancypagestyle{customheader}{
    \fancyhf{} 
    \fancyhead[C]{\raisebox{6ex}[0pt]{\small Proceedings of the Thirty-Second International Joint Conference on Artificial Intelligence (IJCAI-23)}}
}

%% file: settings/paper-status.tex
\newif\ifunderreview
\newif\ifsubmission
\newif\ifappendix

\underreviewfalse

\submissiontrue

\appendixtrue

\ifsubmission
\newcommand{\todo}[1]{}
\newcommand{\replace}[2]{}
\newcommand{\sw}[1]{}

\else
\newcommand{\todo}[1]{\textbf{\textcolor{red}{[TODO: #1]}}}

\newcommand{\replace}[2]{\textbf{\textcolor{red}{[del: \cancel{#1}]}}\textbf{\textcolor{blue}{[new: #2]}}}

\newcommand{\sw}[1]{\textbf{\textcolor{cyan}{[SW: #1]}}}

\usepackage[inline]{showlabels}

\fi

%% file: settings/user-commands.tex
\makeatletter
\newcommand{\customlabel}[2]{%
   \protected@write \@auxout {}{\string \newlabel {#1}{{#2}{\thepage}{#2}{#1}{}} }%
   \hypertarget{#1}{}
}
\makeatother

\newcommand{\argmin}{\mathop{\mathrm{argmin}}}
\newcommand{\argmax}{\mathop{\mathrm{argmax}}}

\newcommand{\D}{\mathcal{D}}
\newcommand{\Dg}{\mathcal{D}^{(g)}}
\newcommand{\Dl}{\mathcal{D}^{(l)}}
\newcommand{\cistar}[1]{c_{#1}^{\star}}
\newcommand{\gcistar}[1]{{\gamma_{c_{#1}^{\star}}}}

\newcommand{\xv}{\boldsymbol{x}}

\newcommand{\Xv}{\mathcal{X}}
\newcommand{\cv}{\boldsymbol{c}}

\newcommand{\truegamma}{\gamma_{i}^{\mathrm{true}}}

\newcommand{\xopt}{\xv_{\mathrm{opt}}}
\newcommand{\pd}[2]{\frac{\partial #1}{\partial #2}}
\newcommand{\eci}{\mathrm{ECI}_{f^{\star}}[\xv | \cv^{\star}, \D]}
\newcommand{\EI}{\mathrm{EI}}
\newcommand{\prob}{\mathbb{P}}
\newcommand{\rank}{\stackrel{\mathrm{rank}}{\simeq}}
\newcommand{\rel}{r^{\mathrm{rel}}}

\newcommand{\citewithname}[2]{#1 \textit{et al.} \shortcite{#2}}

%% file: settings/author-info.tex
\title{c-TPE: Tree-Structured Parzen Estimator with Inequality Constraints \\ for Expensive Hyperparameter Optimization}

\ifunderreview

\author{
  Anonymous authors
\affiliations
Paper under double-bline review
}

\else

\author{
  Shuhei Watanabe
  \And
  Frank Hutter
\affiliations
Department of Computer Science, University of Freiburg, Germany\\
\emails
\{watanabs, fh\}@cs.uni-freiburg.de
}

\fi

\maketitle

%% file: sections/abstract/main.tex
\begin{abstract}
  Hyperparameter optimization (HPO) is crucial for strong performance of deep learning algorithms
  and real-world applications often impose some constraints, such as on memory usage or latency, on top of the performance requirement.
  In this work, we propose constrained TPE (c-TPE), an extension of the widely-used
  versatile Bayesian optimization method,
  tree-structured Parzen estimator (TPE),
  to handle these constraints.
  Our proposed extension goes beyond a simple combination of an existing acquisition function and the original TPE,
  and instead includes modifications that address issues that cause poor performance.
  We thoroughly analyze these modifications both empirically and theoretically,
  providing insights into how they effectively overcome these challenges.
  In the experiments, we demonstrate that c-TPE exhibits
  the best average rank performance among existing methods
  with statistical significance on $81$ expensive HPO problems with inequality constraints.
  Due to the lack of baselines, we only discuss the applicability of our method to hard-constrained optimization in Appendix~\ref{appx:hard-constraints:section}.
  \textcolor{magenta}{The implementation is now available via OptunaHub~\cite{ozaki2025optunahub}.}
\end{abstract}

%% file: sections/introduction/main.tex
\section{Introduction}
\label{introduction-section}
While deep learning (DL) has achieved various breakthrough successes,
its performance highly depends on the proper settings of
its hyperparameters~\cite{chen2018bayesian,melis2018on}.
Furthermore, practical applications often impose
several constraints on memory usage or latency of inference,
making it necessary to apply constrained hyperparameter optimization (HPO).

Recent developments in constrained HPO have led to
the emergence of
new acquisition functions (AFs)~\cite{gardner2014bayesian,hernandez2015predictive,eriksson2021scalable}
in Bayesian optimization (BO) with Gaussian process (GP),
which judge the promise of a configuration based on the surrogate model.
While GP-based methods offer theoretical advantages,
recent open source softwares (OSS) for HPO, such as
Optuna~\cite{akiba2019optuna}, Hyperopt~\cite{bergstra2013making},
and Ray~\cite{liaw2018tune}, instead employ the tree-structured Parzen estimator
(TPE)~\cite{bergstra2011algorithms,bergstra2013making,watanabe2023tpe},
a variant of BO using the density ratio of kernel density estimators for good and bad observations, as the main algorithm, and
Optuna played a pivotal role in HPO of DL models in winning Kaggle competitions~\cite{open-images-2019-object-detection,happy-whale-and-dolphin}.
Despite its versatility for expensive HPO problems, the existing AFs are not directly applicable to TPE
and no study has been conducted on TPE's extension to constrained optimization.

In this paper,
we propose c-TPE, a constrained optimization method that generalizes TPE.
We first show that it is possible to integrate the original TPE into the existing AF proposed by \citewithname{Gelbart}{gelbart2014bayesian}, which uses the product of AFs for the objective and each constraint, and thus TPE can be generalized to constrained settings.
Then, a na\"ive extension, which calculates AF
by the product of density ratios for the objective and each constraint with the same split algorithm,
could be simply obtained;
however, the na\"ive extension suffers from performance degradation under some circumstances.
To circumvent these pitfalls,
we propose
(1) the split algorithm that includes a certain number of feasible solutions,
and (2) an AF defined by the product of relative density ratios, and
analyze their effects empirically and theoretically.

In the experiments,
we demonstrate (1) the strong performance of c-TPE with statistical significance
on expensive HPO problems
and (2) robustness to changes in the constraint level.
Notice that we briefly discuss the applicability of our method to hard-constrained optimization in Appendix~\ref{appx:hard-constraints:section}, and the limitations of our work in Appendix~\ref{section:suppl:limitations}, which stem from our choices of search spaces being limited to tabular benchmarks to enable the stability analysis of the performance variations depending on constraint levels.

In summary, the main contributions of this paper are to:
\begin{enumerate}
  \vspace{-1mm}
  \item prove that TPE can be extended to
  constrained settings using the AF proposed by \citewithname{Gelbart}{gelbart2014bayesian},
  \vspace{-1mm}
  \item present two pitfalls in the na\"ive extension
  and describe how our modifications mitigate those issues,
  \vspace{-1mm}
  \item provide the stability analysis of the performance variations depending on constraint levels, and
  \vspace{-1mm}
  \item demonstrate that the proposed method outperforms existing methods
  with statistical significance
  on average on $9$ tabular benchmarks with $27$ different settings.
  \vspace{-1mm}
\end{enumerate}
The implementation and the experiment scripts are available at
\ifunderreview
\url{https://anonymous.4open.science/r/constrained-tpe-342C/}.
\else
\url{https://github.com/nabenabe0928/constrained-tpe/}.
\fi

%% file: sections/background/main.tex
\vspace{0mm}
\section{Background}
\label{background-section}

\input{sections/background/bo.tex}

\input{sections/background/tpe.tex}

\input{sections/background/cbo.tex}

%% file: sections/background/bo.tex
\vspace{-1mm}
\subsection{Bayesian Optimization (BO)}
Suppose we would like to \textbf{minimize} a validation loss
metric
$f(\xv) = \mathcal{L}(\xv, \mathcal{A}, \D_{\rm train}, \D_{\rm val})$
of a supervised learning algorithm $\mathcal{A}$ given training
and validation datasets $\D_{\rm train}, \D_{\rm val}$,
then the HPO problem is defined as follows:
\begin{equation}
  \begin{aligned}
    \xopt \in
    \argmin_{\xv \in \Xv}
    f(\xv).
  \end{aligned}
\end{equation}
Note that $\xv \in \Xv$ is a hyperparameter configuration,
$\Xv = \Xv_1 \times \dots \times \Xv_D$
is the search space of the hyperparameter configurations,
and $\Xv_d\subseteq  \mathbb{R}~(\text{for }d = 1,\dots,D)$ is the domain of
the $d$-th hyperparameter.
In Bayesian optimization (BO)~\cite{brochu-arXiv10a,shahriari-ieee16a,garnett-bayesian22},
we assume that $f(\xv)$
is expensive and perform the optimization
in a surrogate space given a set of observations $\D \coloneqq \{(\xv_n, f_n)\}_{n=1}^N$.
In each iteration of BO, we build a predictive model $p(f|\xv, \D)$
and optimize an AF to yield the next configuration.
A common AF choice is
the following expected improvement (EI)~\cite{jones1998efficient}:
\begin{equation}
  \begin{aligned}
    \EI_{f^\star}[\xv | \D] =
    \int_{-\infty}^{f^\star}
    (f^\star - f)p(f | \xv, \D) df.
  \end{aligned}
  \label{expected-improvement-eq}
\end{equation}
Another common choice is the following probability of improvement (PI)~\cite{kushner1964new}:
\begin{equation}
  \begin{aligned}
    \prob[f \leq f^\star | \xv, \D] =
    \int_{-\infty}^{f^\star} p(f | \xv, \D) df.
  \end{aligned}
\end{equation}

%% file: sections/background/tpe.tex
\vspace{-2mm}
\subsection{Tree-Structured Parzen Estimator (TPE)}
\label{tpe-subsection}
TPE~\cite{bergstra2011algorithms,bergstra2013making}
is a variant of BO and it uses EI.
See Watanabe~\shortcite{watanabe2023tpe} to better understand the algorithm components.
To transform Eq.~(\ref{expected-improvement-eq}),
we assume the following:
\begin{eqnarray}
  p(\xv | f, \D) = \left\{
  \begin{array}{ll}
    p(\xv | \Dl) & (f \leq f^\gamma) \\
    p(\xv | \Dg) & (f > f^\gamma)
  \end{array}
  \right.
  \label{eq:l-and-g-trick}
\end{eqnarray}
where $\Dl, \Dg$ are the observations
with $f_n \leq f^\gamma$ and $f_n > f^\gamma$, respectively.
Note that 
$f^\gamma$ is the top-$\gamma$ quantile objective value in $\D$ at each iteration
and 
$p(\xv | \Dl), p(\xv | \Dg)$
are built by
the kernel density estimator~\cite{bergstra2011algorithms,bergstra2013making,falkner2018bohb}.
Combining Eqs.~(\ref{expected-improvement-eq}), (\ref{eq:l-and-g-trick})
and Bayes' theorem,
the AF of TPE is computed as \cite{bergstra2011algorithms}:
\begin{equation}
  \begin{aligned}
    \EI_{f^\star}[\xv | \D]
     & \rank r(\xv |\D) \coloneqq p(\xv | \Dl) / p(\xv | \Dg)
  \end{aligned}
  \label{tpe-transformation}
\end{equation}
where $\phi(\xv) \rank \psi(\xv)$ implies
that the functions are order isomorphic and
$\forall \xv, \xv^\prime \in \Xv, \phi(\xv) \leq \phi(\xv^\prime) \Leftrightarrow \psi(\xv) \leq \psi(\xv^\prime)$
holds
and we use $f^\star = f^\gamma$ at each iteration.
In each iteration, TPE samples configurations from
$p(\xv | \Dl)$ and takes the configuration
that achieves the maximum $r(\xv|\D)$.

%% file: sections/background/cbo.tex
\subsection{\normalsize{Bayesian Optimization with Unknown Constraints}}
\label{cbo-subsection}
We consider unknown constraints
$c_i(\xv) = \mathcal{C}_i(\xv, \mathcal{A}, \D_{\rm train}, \D_{\rm val})$,
e.g. memory usage of the algorithm $\mathcal{A}$ given
the configuration $\xv$.
Then the optimization is formulated as follows:
\begin{equation}
  \begin{gathered}
    \xopt \in
    \argmin_{\xv \in \Xv}
    f(\xv) \\
    \text{subject to } \forall i \in \{1,\dots,C\},
    c_i(\xv) \leq \cistar{i}
    \label{eq:constraint-optimization-formula}
  \end{gathered}
\end{equation}
where $\cistar{i} \in \mathbb{R}$ is a threshold
for the $i$-th constraint.
Note that we reverse the sign of the inequality if constraints must be larger than a given threshold.
To extend BO to constrained optimization,
the following expected constraint improvement (ECI) has been proposed \cite{gelbart2014bayesian}:
\begin{equation}
  \begin{aligned}
    \eci =
    \EI_{f^\star}[\xv | \D]
    \prob(c_1 \leq \cistar{1}, \dots, c_C \leq \cistar{C} | \xv, \D).
  \end{aligned}
  \label{expected-constraint-general}
\end{equation}
where $\cv^\star = [\cistar{1},\dots,\cistar{C}] \in \mathbb{R}^C$
and $\D = \{(\xv_n, f_n, \cv_n)\}_{n=1}^N$ is a set of observations,
and $\cv_n = [c_{1,n}, \dots, c_{C,n}] \in \mathbb{R}^C$ is the $n$-th observation of each constraint.
However, the following simplified factorized form is the common choice:
\begin{equation}
  \begin{aligned}
    \eci =
    \EI_{f^\star}[\xv | \D]
    \prod_{i=1}^C \prob(c_i \leq \cistar{i} | \xv, \D),
  \end{aligned}
  \label{eq:expected-constraint-simple}
\end{equation}
Since few methods are available for hard-constrained optimization, we only discuss the applicability of our method to this setting in Appendix~\ref{appx:hard-constraints:section}.

%% file: sections/methods/main.tex
\section{Constrained TPE (c-TPE)}
\label{main:section:methods}
In this section, we first prove that
TPE can be extended to constrained settings
via the simple product of AFs.
Then we describe an extension na\"ively inspired by the original TPE
and discuss two pitfalls hindering efficient search.
Finally, we present modifications for those pitfalls
and analyze the effects on toy problems.

Note that throughout this paper,
we define the term
\textit{$\gamma$-quantile value} $f^\gamma$
as the top-$\gamma$ quantile function value,
$\gcistar{}$ as the quantile of $c^\star$,
and \textit{$\Gamma$-feasible domain}
as the feasible domain in the search space $\Xv$ that covers $100 \times \Gamma \%$ of $\Xv$.
For the formal definitions, see Appendix~\ref{appx:subsection:preliminaries}.
Furthermore, we consider two assumptions mentioned in Appendix~\ref{appx:section:assumptions}
and those assumptions allow the whole discussion to be extended to search spaces
with categorical parameters.

\input{sections/methods/acq.tex}

\input{sections/methods/two-pitfalls.tex}

%% file: sections/methods/acq.tex
\subsection{Na\"ive Acquisition Function}
\label{main:acq-func:section:acq-func}
Suppose we would like to solve constrained optimization problems
formalized in Eq.~(\ref{eq:constraint-optimization-formula})
with ECI.
To realize ECI in TPE,
we first show the following proposition.
\begin{proposition}
  $\EI_{f^\star}[\xv \mid \D] \propto \prob(f \leq f^\star \mid \xv, \D)$ holds
  under the $\mathrm{TPE}$ formulation.
  \label{thm:ei-vs-pi-is-constant}
\end{proposition}
The proof is provided in Appendix~\ref{appx:proof:section:pi-is-ei}.
Since PI and EI
are equivalent under the TPE formulation,
we obtain the following by
combining Proposition~\ref{thm:ei-vs-pi-is-constant} and
Eq.~(\ref{eq:expected-constraint-simple}):
\begin{equation}
  \begin{aligned}
    \eci &\propto \underbrace{
      \prob(f \leq f^\star | \xv, \D)
    }_{\rank r_0(\xv | \D)}
    \prod_{i=1}^C
    \underbrace{
      \prob(c_i \leq \cistar{i} | \xv, \D)
    }_{\rank r_i(\xv | \D)}.
  \end{aligned}
  \label{main:method:eq:eci-transformation-derivation}
\end{equation}
Note that we provide the definition of $r_{i}(\xv | \D)$
for $i \in \{0, 1, \dots, C\}$ in the next section.

%% file: sections/methods/two-pitfalls.tex
\subsection{Two Pitfalls in Na\"ive Extension}

\input{sections/methods/naive-extension.tex}

\input{sections/methods/pitfall-issue01.tex}

\input{sections/methods/pitfall-issue02.tex}

%% file: sections/methods/naive-extension.tex
\input{sections/methods/c-tpe-algo.tex}

\subsubsection{Na\"ive Extension and Modifications}
\label{main:methods:section:naive-extension}
From the discussion above, we could na\"ively extend
the original TPE to constrained settings using
the split in Eq.~(\ref{eq:l-and-g-trick})
and the AF in Eq.~(\ref{tpe-transformation}).
More specifically, the na\"ive extension computes the AF as follows:
\begin{enumerate}
  \item Pick the $\lceil\gamma|\D|\rceil$-th best objective value $f^\star$ in $\D$,
  \vspace{-1mm}
  \item Split $\D$ into $\Dl_0$ and $\Dg_0$ at $f^\star$,
  and $\D$ into $\Dl_i$ and $\Dg_i$ at $\cistar{i}$ for $i \in \{1,\dots, C\}$,
  \vspace{-1mm}
  \item Build kernel density estimators $p(\xv|\Dl_i)$, $p(\xv|\Dg_i)$ for 
  $i \in \{0,\dots, C\}$, and
  \vspace{-1mm}
  \item Take the product of density ratios $\prod_{i=0}^C r_i(\xv|\D) \coloneqq \prod_{i=0}^C p(\xv|\Dl_i)/p(\xv|\Dg_i)$
  as the AF.
\end{enumerate}
Note that as $\cistar{i}$ is a user-defined threshold, $\cistar{i}$ is fixed during the optimization.
Although this implementation could be naturally inspired
by the original TPE, Operations 1 and 4 could incur
performance degradation under
(1) small overlaps in top domains for the objective
and feasible domains, or
(2) vanished constraints.

For this reason, we change Operations 1 and 4 as follows:
\begin{itemize}
  \item Pick the $\lceil \gamma|\D|\rceil$-th best \textbf{feasible} objective value $f^\star$ in $\D$~(Line~\ref{line:constraint-split}), and
  \vspace{-1mm}
  \item Take the product of \textbf{relative} density ratios
  $\prod_{i=0}^C \rel_i(\xv|\D)
  \coloneqq \prod_{i=0}^C 
  (\hat{\gamma}_i +
  (1 - \hat{\gamma}_i)
  r_i(\xv | \D)^{-1})^{-1}
  $ as the AF~(Line~\ref{line:eci-calculation}).
\end{itemize}
Note that we color-coded the modifications in Algorithm~\ref{alg:c-tpe-algo} and
we define $\hat{\gamma}_i \coloneqq |\Dl_i|/|\D|$.
Intuitively, when all configurations satisfy the $i$-th constraint,
i.e., $|\Dl_i| = |\D| \Rightarrow \hat{\gamma}_i = 1$,
we trivially yield $\rel_i = 1$;
therefore, the $i$-th constraint will be ignored and it is equivalent to $\forall \xv \in \Xv, \prob[c_i \leq \cistar~|\xv, \D] = 1$. 
Additionally, the following corollary guarantees the mathematical validity of our algorithm:
\begin{corollary}
  $\eci \propto \prod_{i=0}^C \rel_i(\xv|\D)$ under the $\mathrm{TPE}$ formulation. 
  \label{main:method:proof:eci-is-product-of-relative-density-ratios}
\end{corollary}
We provide the proof in Appendix~\ref{appx:proof:section:eci-is-product-of-relative-density-ratios}.

The split algorithm in the original TPE by \citewithname{Bergstra}{bergstra2013making}
first sorts the observations $\D$ by $f$
and takes the first $\lceil\sqrt{N} / 4\rceil$ observations
as $\Dl_0$ and the rest as $\Dg_0$.
On the other hand,
our method includes all the observations until the
$\lceil\sqrt{N} / 4\rceil$-th \textbf{feasible} observation
into $\Dl_0$
and the rest into $\Dg_0$. This split algorithm matches the original algorithm when $\Gamma = 1$.
For the split of constraints,
we first check the upper bound of $\{c_{i,n}\}_{n=1}^{N}$
that satisfies a given threshold $\cistar{i}$ and let this value be $c_i^\prime$.
Note that $c_{i,n}$ is the $i$-th constraint value in the $n$-th observation.
If such a value does not exist, we take the best value
$\min\{c_{i,n}\}_{n=1}^{N}$ so that
the optimization of this constraint will be strengthened~(see Theorem~\ref{main:vanished-constraints:thm:tight-constraint-have-more-priority}).
Then we split $\D$ into $\Dl_i$ and $\Dg_i$
so that $\Dl_i$ includes only observations that satisfy
$c_{i, n} \leq c_i^\prime$ and vice versa.
We describe more details in Appendix~\ref{suppl:section:split-criterion} and the applicability to hard-constrained optimization in Appendix~\ref{appx:hard-constraints:section}.
We discuss why these modifications mitigate the issues in the next section.

%% file: sections/methods/c-tpe-algo.tex
\begin{algorithm}[tb]
  \caption{c-TPE algorithm (With modifications)}
  \label{alg:c-tpe-algo}
  \begin{algorithmic}[1]
    \State{$N_{\mathrm{init}}$ (The number of initial configurations), $N_s$ (The number of candidates to consider in the optimization of the AF)}
    \State{$\D \leftarrow \emptyset$}
    \For{$n = 1, \dots, N_{\mathrm{init}}$}
    \State{Randomly pick $\xv$}
    \State{$\D \leftarrow \D \cup \{(\xv, f(\xv), c_1(\xv), \dots, c_C(\xv))\}$}
    \EndFor
    \While{Budget is left}
    \State{$\mathcal{S} = \emptyset$}
    \For{$i = 0, \dots, C$}
    \State{\textcolor{cyan}{Split $\D$ into $\Dl_i$ and $\Dg_i$}, $\hat{\gamma}_i \leftarrow |\Dl_i| / |\D|$}
    \label{line:constraint-split}
    \State{Build $p(\cdot | \Dl_i), p(\cdot | \Dg_i)$}
    \State{$\{\xv_j\}_{j=1}^{N_s} \sim p(\cdot | \Dl_i), \mathcal{S} \leftarrow \mathcal{S} \cup \{\xv_j\}_{j=1}^{N_s}$}
    \EndFor
    \LineComment{See Appendix~\ref{appx:hard-constraints:section} for the hard-constrained version}
    \State{\textcolor{cyan}{Pick $\xopt \in \argmax_{\xv \in \mathcal{S}} \prod_{i=0}^C \rel_i(\xv | \D)$}}
    \label{line:eci-calculation}
    \State{$\D \leftarrow \D \cup \{(\xopt, f(\xopt), c_1(\xopt), \dots, c_C(\xopt))\}$}
    \EndWhile
  \end{algorithmic}
\end{algorithm}

%% file: sections/methods/pitfall-issue01.tex
\subsubsection{Issue I: Vanished Constraints}
\label{main:pitfall-issue01:section:issue01}

We refer to
constraints that are satisfied in almost all configurations
as \emph{vanished constraints}.
In other words, if the $i$-th constraint $c_i$ is a vanished constraint,
its quantile is $\hat{\gamma}_i \coloneqq \hat{\gamma}_{\cistar{i}} \simeq 1$.
In this case, $r_i(\xv|\D)$ should be a constant value
as $\prob(c_i\leq \cistar{i}|\xv,\D) = 1$ holds for almost all configurations $\xv$.
As discussed in Section~\ref{main:methods:section:naive-extension}, the relative density ratio $\rel_i(\xv|\D)$ resolves this issue
and it can be written more formally as follows:
\begin{corollary}
  Assuming the feasible domain quantile $\Gamma = 1$,
  then $\prod_{i=0}^C \rel_i(\xv|\D) \rank r_0(\xv|\D)$ holds.
  \label{thm:ctpe-converges-to-single-objective}
\end{corollary}
Recall that we previously defined
$r_0(\xv|\D) \coloneqq p(\xv|\Dl_0) / p(\xv|\Dg_0)$
for $\Dl_0, \Dg_0$ obtained by splitting $\D$ at $f^\star$.
The proof is provided in Appendix~\ref{appx:proof:section:c-tpe-is-tpe-for-vanished-constraints}.
Corollary~\ref{thm:ctpe-converges-to-single-objective} indicates that the AF of c-TPE is equivalent to that of the original TPE when $\Gamma = 1$ and it means that our formulation achieves $\prob(c_i\leq \cistar{i}|\xv,\D) = 1$ if $\hat{\gamma}_i = 1$.
Corollary~\ref{thm:ctpe-converges-to-single-objective} is a special case of the following theorem:
\begin{theorem}
  Given a pair of constraint thresholds $\cistar{i}, \cistar{j}$
  and the corresponding quantiles $\hat{\gamma}_i,\hat{\gamma}_j (\hat{\gamma}_i \leq \hat{\gamma}_j)$,
  if $r_i + \frac{\hat{\gamma}_i}{1 - \hat{\gamma}_i}r_i^2 \leq r_j + \frac{\hat{\gamma}_j}{1 - \hat{\gamma}_j}r_j^2$ 
  holds,
  then
  \begin{equation}
  \begin{aligned}
    \pd{\prod_{k=0}^C \rel_k(\xv|\D)}{r_i} \geq \pd{\prod_{k=0}^C \rel_k(\xv|\D)}{r_j} \geq 0
  \end{aligned}
  \end{equation}
  holds where the first equality holds
  if $\hat{\gamma}_i = \hat{\gamma}_j$ and $r_i = r_j$
  and the second one holds iff $\hat{\gamma}_j = 1$.
  \label{main:vanished-constraints:thm:tight-constraint-have-more-priority}
\end{theorem}
The proof is provided in Appendix~\ref{appx:proof:section:tighter-constraint-has-more-priority}.
Roughly speaking, Theorem~\ref{main:vanished-constraints:thm:tight-constraint-have-more-priority} implies that
our modified AF puts more priority on the variations of the density ratios
with lower quantiles, i.e., $r_i$ in the statement above, when $r_i = r_j$.

\begin{figure}
  \centering
  \includegraphics[width=0.49\textwidth]{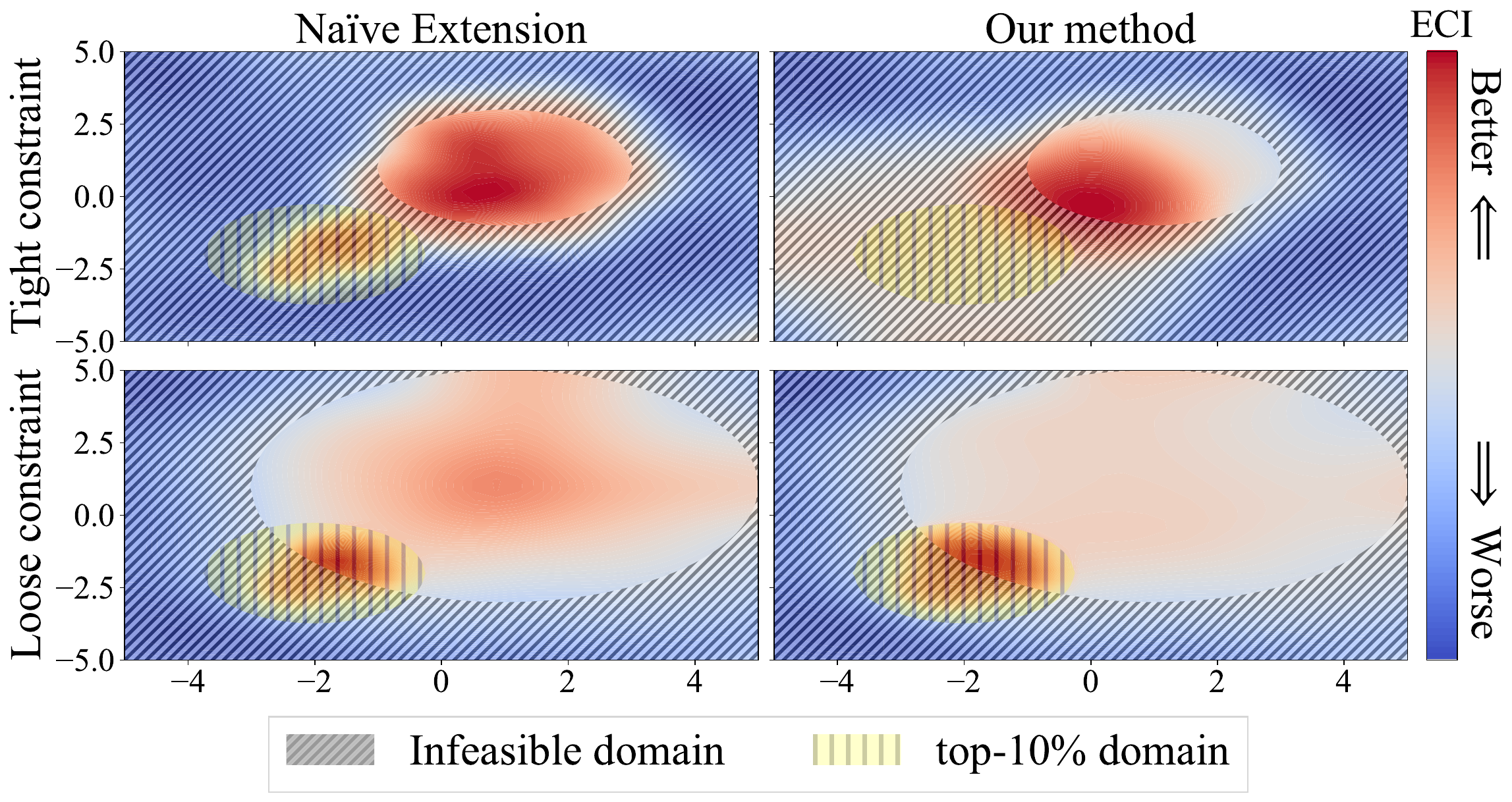}
  \vspace{-7mm}
  \caption{
    Heat maps of the AF in the na\"ive extension (\textbf{Left column})
    and our c-TPE (\textbf{Right column})
    with a tight (\textbf{Top row}, $c_1 = 4$) or loose (\textbf{Bottom row}, $c_1 = 16$) constraint.
    For fair comparisons, we use a fixed set of $200$ randomly sampled configurations
    to compute the AF for all settings.
    In principle, red regions
    have higher AF values
    and the next configuration is likely to be picked from here.
  }
  \vspace{-1mm}
  \label{main:issue01:fig:vanished-constraints}
\end{figure}

We empirically and intuitively present the effect of
Theorem~\ref{main:vanished-constraints:thm:tight-constraint-have-more-priority}
in Figure~\ref{main:issue01:fig:vanished-constraints}.
We used the objective function $f(x, y) = (x + 2)^2 + (y + 2)^2$
and the constraint $c_1(x,y) = (x - 1)^2 + (y - 1)^2 \leq \cistar{1} \in \{4, 16\}$
and visualize the heat maps of the AF using
exactly the same observations for each figure.
Note that all used parameters are described in Appendix~\ref{appx:experiment-settings:section:details}.
As mentioned earlier, since the na\"ive extension (\textbf{Left column}) does not decay the contribution from the objective or the constraint with a large $\hat{\gamma}_i$, it has two peaks.
For our algorithm, however,
we only have one peak between the top-$10\%$ domain
and the feasible domain
because our AF decays the contribution
from either the objective or the constraint based
on their quantiles $\hat{\gamma}_i$ as mentioned in
Theorem~\ref{main:vanished-constraints:thm:tight-constraint-have-more-priority}.
More specifically, for the tight constraint case (\textbf{Top right}),
since the feasible domain quantile $\hat{\gamma}_1 \simeq 0.12$
is relatively small compared to the top-solution quantile
$\hat{\gamma}_0 \simeq 0.3$,
the peak in the top-$10\%$ domain vanishes.
Notice that we discuss why we have the peak
not at the center of the feasible domain, but
between the feasible domain and the top-$10\%$ domain
in the next section.
For the loose constraint case (\textbf{Bottom right}),
$\hat{\gamma}_1 \simeq 0.50$ is much larger than
$\hat{\gamma}_0\simeq 0.02$
and this decays the contribution from the center of the feasible domain
where we have the largest $r_1(\xv|\D)$.
As mentioned in Corollary~\ref{thm:ctpe-converges-to-single-objective}, $\rel_i(\xv|\D) = 1$ holds for $i \in \{1,\dots,C\}$ when $\Gamma = 1$, and thus the AF coincides with that for the single objective optimization.
Note that since we obtain $\hat{\gamma}_0 = 1.0$ in the case of all observations being infeasible,
the objective function will be ignored and only constraints will be optimized.

%% file: sections/methods/pitfall-issue02.tex
\subsubsection{\small{Issue II: Small Overlaps in Top and Feasible Domains}}
\label{main:pitfall-issue02:section:issue02}
Since the original TPE algorithm just takes the top-$\gamma$ quantile observations,
it does not guarantee that $\Dl$ has feasible solutions.
We explain its effect using Figure~\ref{main:issue01:fig:vanished-constraints}.
For the tight constraint case (\textbf{Top row}),
an overlap between the feasible domain and the top-$10\%$ domain does not exist
and it causes the two peaks in the AF for the original split algorithm (\textbf{Top left});
however, it is necessary for constrained optimization to sample intensively within feasible domains.
In turn, we modify the split algorithm to include a certain number of feasible solutions.
This modification leads to the large white circle that embraces the top-$10\%$ domain (\textbf{Top right}).
As a result, our algorithm yields a peak at the overlap between the large white circle and the feasible domain.

\begin{figure}[t]
  \centering
  \includegraphics[width=0.49\textwidth]{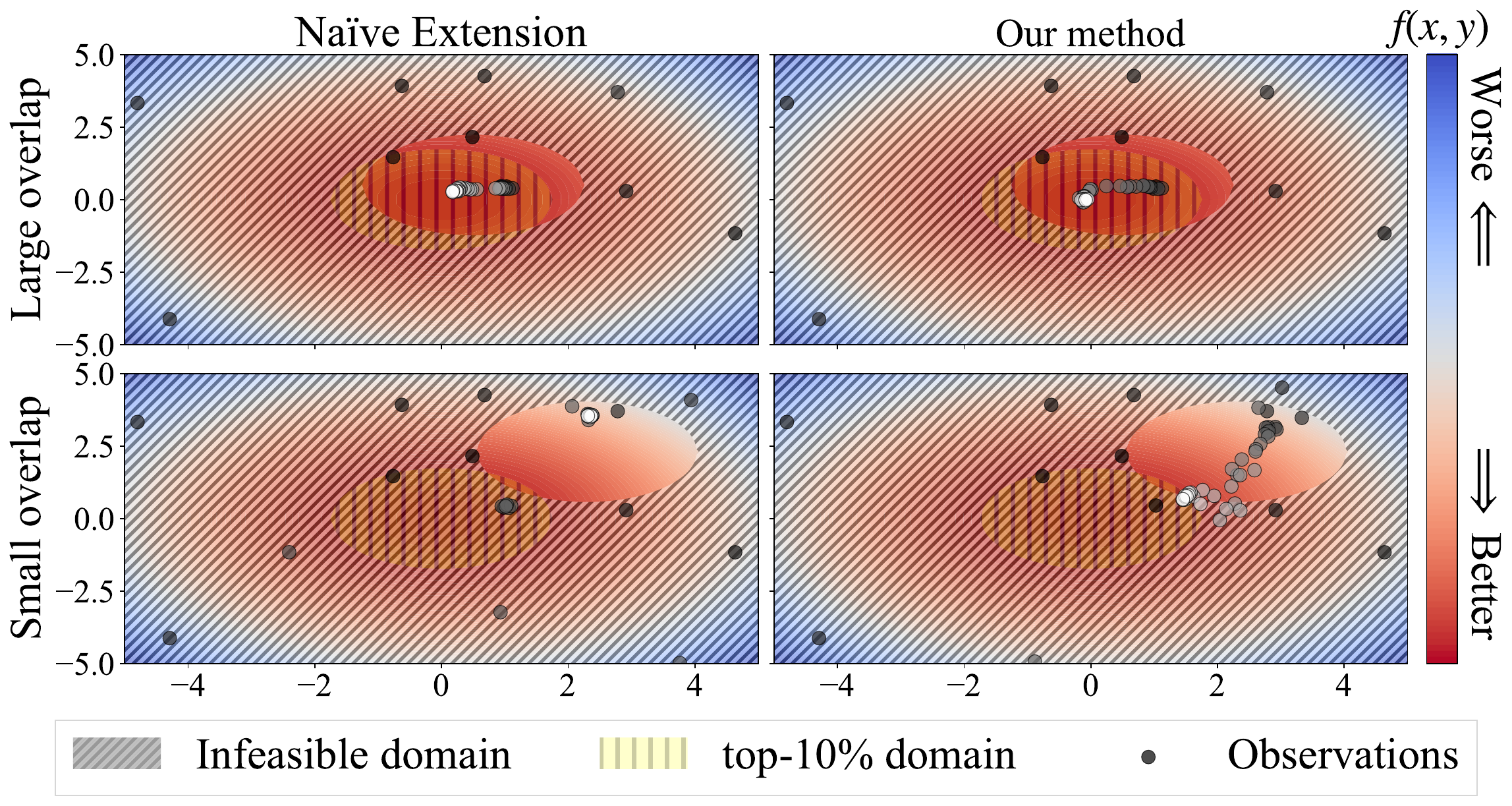}
  \vspace{-7mm}
  \caption{
    Scatter plots of observations obtained by
    the na\"ive extension (\textbf{Left column})
    and our c-TPE (\textbf{Right column})
    on a large (\textbf{Top row}, $z = 0.5$) or small (\textbf{Bottom row}, $z = 2.3$)
    overlap between the top-$10\%$ domain and feasible domain.
    Each figure shows the 2D search space for each task
    and the observations obtained during optimization
    are plotted.
    Earlier observations are colored black
    and later observations are colored white.
    Each figure has $50$ observations.
  }
  \vspace{-1mm}
  \label{main:issue02:fig:small-overlap}
\end{figure}

In Figure~\ref{main:issue02:fig:small-overlap},
we visualize how our algorithm and the na\"ive extension
sample configurations using a toy example.
We used the objective function $f(x, y) = x^2 + y^2$
and the constraint $c_1(x,y) = (x - z)^2 + (y - z)^2 \leq \cistar{1} = 3$,
where $z \in \{0.5, 2.3\}$.
This experiment also follows the settings used in Appendix~\ref{appx:experiment-settings:section:details}
and both algorithms share the initial configurations.
In the large overlap case (\textbf{Top row}), both algorithms search similarly.
In contrast to this case,
the small overlap case (\textbf{Bottom row}) obtained different sampling behaviors.
While our algorithm (\textbf{Bottom right}) samples intensively
at the boundary between the feasible domain and the top-$10\%$ domain,
the na\"ive extension (\textbf{Bottom left}) does not.
Furthermore, we can see a trajectory from
the top right of the feasible domain to the boundary for our algorithm
and it exists only in our algorithm
although both methods have some observations,
which are colored in dark gray, meaning that they were obtained at the early stage of the optimization,
in the top right of the feasible domain.
Based on Figure~\ref{main:issue01:fig:vanished-constraints} (\textbf{Top right}),
we can infer that
this is because we include some feasible solutions in $\Dl_0$
and the peak of the AF will be shifted
toward the top-$10\%$ domain in our algorithm.

%% file: sections/experiments/main.tex

\section{Experiments}
\label{main:section:experiments}

\input{sections/experiments/experiment-settings.tex}

\input{sections/experiments/perf-over-time-in-main.tex}

\input{sections/experiments/robustness-to-feasible-domain-size.tex}

\input{sections/experiments/avg-rank-discussion.tex}

%% file: sections/experiments/experiment-settings.tex
\input{sections/experiments/test-perf-group-by-quantiles-in-main.tex}

\subsection{Setup}
The evaluations were performed on the following $10$ tabular benchmarks:
\begin{enumerate}
  \item HPOlib
  (Slice Localization, Naval Propulsion, Parkinson's Telemonitoring, Protein Structure)~\cite{klein2019tabular}:
  All with $6$ numerical and $3$ categorical parameters;
  \item NAS-Bench-101 (CIFAR10A, CIFAR10B, CIFAR10C)~\cite{ying2019bench}:
  Each with $26$ categorical, $14$ categorical, and $22$ numerical and $5$ categorical parameters, respectively; and
  \item NAS-Bench-201 (ImageNet16-120, CIFAR10, CIFAR100)~\cite{dong2020bench}:
  All with $6$ categorical parameters.
\end{enumerate}
The reason behind this choice is that
tabular benchmarks enable us to control
the quantiles of each constraint $\truegamma$, which
significantly change the feasible domain size and the quality of solutions.
For example, suppose a tabular dataset has $N_{\mathrm{all}}$ configurations
$\{(\xv_n, f_n, \cv_n)\}_{n=1}^{N_{\mathrm{all}}}$
and the dataset is sorted so that it satisfies
$c_{i,1} \leq c_{i,2} \leq \dots \leq c_{i,N_{\mathrm{all}}}$
where $c_{i,n}$ is the $i$-th constraint value in the $n$-th configuration,
then we fix the threshold for the $i$-th constraint $\cistar{i}$ as
$c_{i,\lfloor N_{\mathrm{all}}/10 \rfloor}$ in the setting of $\truegamma=1/10$.
We evaluated each benchmark
with $9$ different quantiles
$\truegamma$ for each constraint
and $3$ different constraint choices.
Constraint choices are network size, runtime, or both.
The search space for each benchmark followed \citewithname{Awad}{awad2021dehb}.

As the baseline methods, we chose:
\begin{enumerate}
  \item \textbf{Random search}~\cite{bergstra2012random},
  \item \textbf{CNSGA-II}~\cite{deb2002fast},
  \footnote{Implementation: \url{https://github.com/optuna/optuna}}
  (population size 8),
  \item \textbf{Noisy ECI (NEI)}~\cite{letham2019constrained}
  \footnote{Implementation: \url{https://github.com/facebook/Ax}},
  \item \textbf{Hypermapper2.0 (HM2)}~\cite{nardi2019practical}
  ~\footnote{Implementation: \url{https://github.com/luinardi/hypermapper}},
  \item \textbf{Vanilla TPE} (optimize only loss as if we do not have constraints), and
  \item \textbf{Na\"ive c-TPE} (the na\"ive extension discussed in Section~\ref{main:section:methods}).
\end{enumerate}
We describe the details of each method and
their control parameters in Appendix~\ref{appx:experiment-settings:section:details}.
Note that all experiments were performed $50$ times with different random seeds
and we evaluated $200$ configurations for each optimization.
Additionally, since the optimizations by NEI and HM2 on CIFAR10C
failed due to the high-dimensional ($22$ dimensions)
continuous search space for NEI and an unknown internal issue for HM2,
we used the results on $9$ benchmarks (other than CIFAR10C) for the statistical test
and the average rank computation.
The results on CIFAR10C by the other methods are
available in Appendix~\ref{section:suppl:robustness-to-percentile}
and
the source code is available at 
\ifunderreview
\url{https://anonymous.4open.science/r/constrained-tpe-342C}
\else
\url{https://github.com/nabenabe0928/constrained-tpe}
\fi
along with complete scripts to reproduce the experiments, tables, and figures.
Queries of c-TPE with $\{50,100,150,200\}$ observations took $\{0.22, 0.24, 0.26, 0.28\}$ seconds, respectively,
for a 30D problem with 8 cores of Core i7-10700.

%% file: sections/experiments/test-perf-group-by-quantiles-in-main.tex
\addtolength{\tabcolsep}{-4pt}
\begin{table*}
  \caption{
    The table shows (Wins/Loses/Ties) of c-TPE against each method
    for optimizations with different constraint levels ($9$ benchmarks $\times$ $3$ constraint choices $=$ $27$ settings).
    The number of wins was counted by comparing medians of performance over $50$ random seeds
    in each setting between two methods.
    Non-bold numbers indicate $p < 0.01$ of
    the hypothesis ``The other method is better than c-TPE''
    by the Wilcoxon signed-rank test.
  }
  \vspace{-2mm}
  \label{tab:rank-group-by-quantiles}
  \makebox[1 \textwidth][c]{       
    \resizebox{0.96 \textwidth}{!}{   
      \begin{tabular}{lcccc|cccc|cccc}
        \toprule
        \multicolumn{1}{c}{Quantiles}
                    & \multicolumn{4}{c}{$\truegamma = 0.1$}
                    & \multicolumn{4}{c}{$\truegamma = 0.5$}
                    & \multicolumn{4}{c}{$\truegamma = 0.9$}
        \\
        \multicolumn{1}{c}{Methods / \# of configs}
                    & 50                                     & 100    & 150    & 200    & 50     & 100    & 150    & 200    & 50               & 100    & 150    & 200
        \\
        \midrule
        Naïve c-TPE & 26/0/1                                 & 27/0/0 & 27/0/0 & 27/0/0 & 25/0/2 & 25/0/2 & 25/1/1 & 25/0/2 & 21/5/1           & 23/1/3 & 21/1/5 & 24/1/2 \\
        Vanilla TPE & 27/0/0                                 & 27/0/0 & 27/0/0 & 27/0/0 & 25/0/2 & 26/0/1 & 26/1/0 & 24/0/3 & \textbf{14/11/2} & 18/8/1 & 15/5/7 & 16/7/4 \\
        Random      & 25/0/2                                 & 26/1/0 & 27/0/0 & 27/0/0 & 27/0/0 & 26/0/1 & 26/0/1 & 27/0/0 & 27/0/0           & 27/0/0 & 27/0/0 & 27/0/0 \\
        CNSGA-II    & 25/0/2                                 & 27/0/0 & 24/0/3 & 24/0/3 & 26/0/1 & 26/0/1 & 26/0/1 & 25/0/2 & 26/1/0           & 27/0/0 & 27/0/0 & 26/0/1 \\
        NEI         & 24/1/2                                 & 27/0/0 & 27/0/0 & 27/0/0 & 27/0/0 & 26/0/1 & 26/0/1 & 27/0/0 & 27/0/0           & 27/0/0 & 27/0/0 & 27/0/0 \\
        HM2         & 23/2/2                                 & 26/1/0 & 25/2/0 & 25/2/0 & 22/3/2 & 23/2/2 & 25/1/1 & 23/0/4 & 27/0/0           & 27/0/0 & 23/0/4 & 26/0/1 \\
        \bottomrule
      \end{tabular}
    }
  }
\end{table*}
\addtolength{\tabcolsep}{4pt}

%% file: sections/experiments/perf-over-time-in-main.tex
\begin{figure*}[t]
  \centering
  \includegraphics[width=0.98\textwidth]{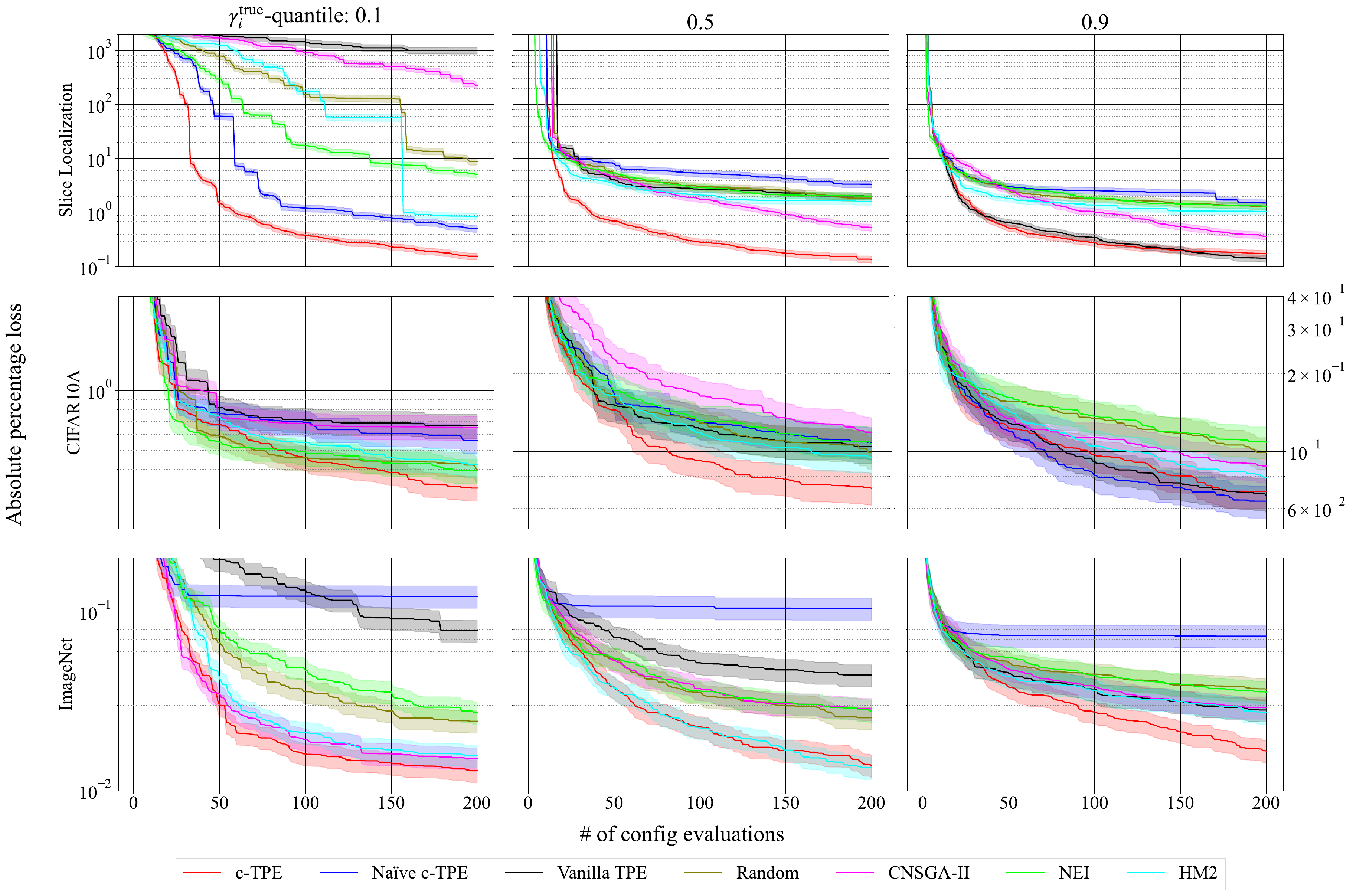}
  \caption{
    The performance curves on Slice Localization in
    HPOlib (\textbf{Top row}),
    CIFAR10A in NAS-Bench-101 (\textbf{Middle row}),
    and ImageNet16-120 in NAS-Bench-201 (\textbf{Bottom row})
    with constraints of runtime and network size.
    We picked
    $\truegamma = 0.1$ (\textbf{Left column}),
    $0.5$ (\textbf{Center column}),
    $0.9$ (\textbf{Right column}).
    The vertical axis shows the absolute percentage loss
    $(f_{\mathrm{observed}} - f_{\mathrm{oracle}}) / f_{\mathrm{oracle}}$
    where $f_{\mathrm{oracle}}$ is determined by looking up
    all feasible configurations in each benchmark.
    Note that each row shares the vertical axis except NAS-Bench-101.
    For $\truegamma = 0.1$ in NAS-Bench-101,
    we separately scaled it for readability.
    Further results are available in
    Appendix~\ref{section:suppl:robustness-to-percentile}.
  }
  \label{fig:main:eval-vs-loss-over-various-constraints}
\end{figure*}

%% file: sections/experiments/robustness-to-feasible-domain-size.tex
\input{sections/experiments/test-perf-group-by-constraints-in-main.tex}
\input{sections/experiments/rank-over-time-in-main.tex}

\subsection{Robustness to Feasible Domain Size}
\label{section:main:experiments:robustness-to-constraint-percentile}
This experiment shows how c-TPE performance
improves given various levels of constraints.
We optimized each benchmark with
the aforementioned three types of constraints and chose
$\truegamma \in \{0.1,0.5,0.9\}$ for each constraint.
All results on other benchmarks are available in Appendix~\ref{section:suppl:robustness-to-percentile}.
Table~\ref{tab:rank-group-by-quantiles} presents
the numbers of wins/loses/ties and statistical significance
by the Wilcoxon signed-rank test and
Figure~\ref{fig:main:eval-vs-loss-over-various-constraints} shows
the performance curves for each benchmark.

As a whole, while the performance of c-TPE is stable across all constraint levels,
that of NEI, HM2, and CNSGA-II varies depending on constraint levels.
Furthermore, Table~\ref{tab:rank-group-by-quantiles} shows that c-TPE is significantly better than other methods in almost all settings.
This experimentally validates the robustness of c-TPE to the variations in constraint levels.

For ImageNet of NAS-Bench-201 (\textbf{Bottom row}),
the na\"ive c-TPE is completely defeated by the other methods
while c-TPE achieves the best or indistinguishable performance from the best.
This gap between c-TPE and the na\"ive c-TPE is caused by the small overlaps discussed in Section~\ref{main:pitfall-issue02:section:issue02}.
For example, only $59\%$ of the top-$10\%$ configurations belong to the feasible domain in NAS-Bench-201 of $\truegamma = 0.9$
although we can usually expect that $90\%$ of them belong to the feasible domain,
and $84\%$ and $77\%$ of those in HPOlib and NAS-Bench-101 actually belong to the feasible domain for $\truegamma = 0.9$, respectively.
The small overlap leads to the performance gap between c-TPE and the vanilla TPE as well.
As TPE is not a uniform sampler and tries to sample from top domains,
$\hat{\gamma}_i$ will not necessarily approach $\truegamma$.
In our case, it is natural to consider $\hat{\gamma}_i$
to be closer to $59\%$ rather than $90\%$
as only $59\%$ of top-$10\%$ configurations are feasible.
As mentioned also in Theorem~\ref{main:vanished-constraints:thm:tight-constraint-have-more-priority},
c-TPE is advantageous to such settings compared to the vanilla TPE and the na\"ive c-TPE.

For CIFAR10A of NAS-Bench-101 (\textbf{Middle row}),
the results show different patterns from the other settings due to the high-dimensional ($D = 26$) nature.
For $\truegamma = 0.1, 0.5$ (\textbf{Left, center}),
most methods exhibit indistinguishable performance
from random search especially in the beginning
because little information on feasible domains is available in the early stages of optimization due to the high dimensionality
although c-TPE outperforms in the end.
In $\truegamma = 0.9$ (\textbf{Right}),
the na\"ive c-TPE is slightly better than
c-TPE due to large overlaps ($84\%$ of the top-$10\%$ configurations are feasible).
It implies that if the search space is high-dimensional
and overlaps in top domains and feasible domains are large,
it might be better to greedily optimize only the objective
rather than regularizing the optimization of the objective as in our modification.

For Slice Localization of HPOlib (\textbf{Top row}), c-TPE outperforms the other methods.
Furthermore, its performance almost coincides with that of the vanilla TPE in $\truegamma = 0.9$ and
it implies that our method gradually decays the priority of each constraint as $\truegamma$ becomes larger.
In fact, the na\"ive c-TPE does not exhibit stability when the constraint level changes
as it does not consider the priority of each constraint and the objective.
This result empirically validates Theorem~\ref{main:vanished-constraints:thm:tight-constraint-have-more-priority}.

%% file: sections/experiments/test-perf-group-by-constraints-in-main.tex
\addtolength{\tabcolsep}{-4pt}
\begin{table*}
  \caption{
    The table shows (Wins/Loses/Ties) of c-TPE against each method
    for optimizations with different constraints ($9$ benchmarks $\times$ $9$ quantiles $=$ $81$ settings).
    The number of wins was counted by comparing medians of performance over $50$ random seeds
    in each setting between two methods.
    In this table, all results indicate $p < 0.01$ of
    the hypothesis ``The other method is better than c-TPE''
    by the Wilcoxon signed-rank test.
  }
  \vspace{-2mm}
  \label{tab:rank-group-by-constraints}
  \makebox[1 \textwidth][c]{       
    \resizebox{0.96 \textwidth}{!}{   
      \begin{tabular}{lcccc|cccc|cccc}
        \toprule
        \multicolumn{1}{c}{Constraints}
                    & \multicolumn{4}{c}{Runtime \& Network size}
                    & \multicolumn{4}{c}{Network size}
                    & \multicolumn{4}{c}{Runtime}
        \\
        \multicolumn{1}{c}{Methods / \# of configs}
                    & 50                                          & 100             & 150             & 200             & 50               & 100             & 150             & 200             & 50               & 100              & 150              & 200
        \\
        \midrule
        Naïve c-TPE & 77/3/1                             & 79/0/2 & 78/0/3 & 79/0/2 & 75/4/2  & 77/1/3 & 76/1/4 & 80/0/1 & 66/8/7  & 71/5/5  & 70/3/8  & 69/2/10 \\
        Vanilla TPE & 73/7/1                             & 75/5/1 & 72/3/6 & 73/4/4 & 69/10/2 & 74/6/1 & 74/3/4 & 72/6/3 & 62/12/7 & 67/10/4 & 62/6/13 & 60/9/12 \\
        Random      & 80/0/1                             & 81/0/0 & 81/0/0 & 80/0/1 & 80/0/1  & 79/2/0 & 80/1/0 & 81/0/0 & 80/0/1  & 78/0/3  & 79/0/2  & 81/0/0  \\
        CNSGA-II    & 80/0/1                             & 79/0/2 & 76/1/4 & 75/2/4 & 77/3/1  & 78/1/2 & 75/2/4 & 75/1/5 & 74/1/6  & 76/0/5  & 74/0/7  & 74/0/7  \\
        NEI         & 79/1/1                             & 81/0/0 & 81/0/0 & 81/0/0 & 79/1/1  & 80/1/0 & 80/1/0 & 81/0/0 & 77/0/4  & 78/0/3  & 79/0/2  & 81/0/0  \\
        HM2         & 74/5/2                             & 77/3/1 & 77/1/3 & 76/2/3 & 76/4/1  & 78/2/1 & 76/2/3 & 78/0/3 & 71/4/6  & 73/2/6  & 67/3/11 & 70/2/9  \\
        \bottomrule
      \end{tabular}
    }
  }
\end{table*}
\addtolength{\tabcolsep}{4pt}

%% file: sections/experiments/rank-over-time-in-main.tex
\begin{figure*}[t]
  \centering
  \includegraphics[width=0.98\textwidth]{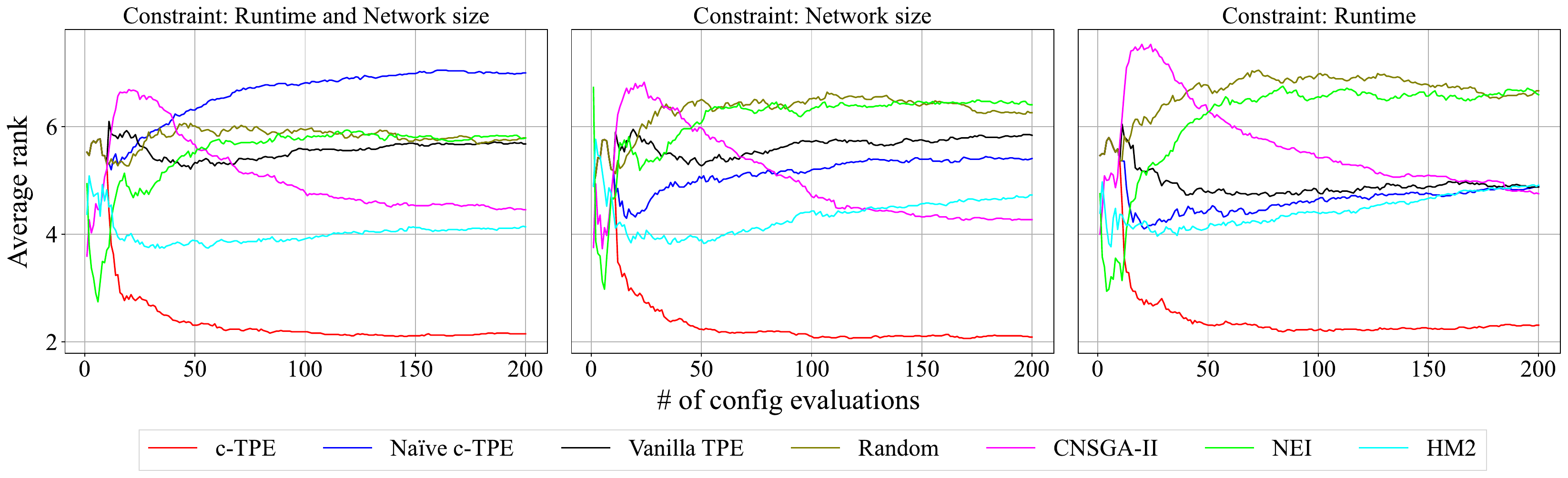}
  \vspace{-3mm}
  \caption{The average rank of each method over the number of evaluations.
  The horizontal axis shows the number of evaluated configurations in optimizations
    and the vertical axis shows the average rank over
    81 settings.
    The title of each figure shows the constraint that the optimizations 
    handled.
  }
  \vspace{-3mm}
  \label{fig:rank-over-time-all}
\end{figure*}

%% file: sections/experiments/avg-rank-discussion.tex
\subsection{Average Rank over Number of Evaluations}
\label{section:main:experiments:rank-over-time}
This experiment demonstrates how c-TPE performance improves
compared to the other methods over the number of evaluations.
Table~\ref{tab:rank-group-by-constraints} presents
the numbers of wins/loses/ties and statistical significance by the Wilcoxon signed-rank test
and Figure~\ref{fig:rank-over-time-all} shows the average rank over $81$ settings ($9$ benchmarks $\times$ $9$ quantiles).

According to Figure~\ref{fig:rank-over-time-all},
c-TPE quickly reaches the top rank and maintains it until the end.
From the figures, we can see that CNSGA-II improves in rank
as the number of evaluations grows.
In fact, since such a slow start is often the case for evolutionary algorithms such as CMA-ES~\cite{loshchilov2016cma}, the quick convergence achieved by c-TPE is appealing.
For the multiple-constraint setting (\textbf{Right}),
while the na\"ive c-TPE is worse than random search due to the small overlap,
c-TPE overcomes this problem as discussed in Section~\ref{main:pitfall-issue02:section:issue02}.
Table~\ref{tab:rank-group-by-constraints} confirmed the anytime performance of c-TPE by the statistical test over all the settings.
All results on individual settings and
quantile-wise average rank are available in
Appendices~\ref{section:suppl:robustness-to-percentile} and
\ref{section:suppl:rank-over-time}.

%% file: sections/related-work/main.tex
\section{Related Work \& Discussion}
\label{main:section:related-works}
ECI was 
introduced by \citewithname{Gardner}{gardner2014bayesian}
and \citewithname{Gelbart}{gelbart2014bayesian}.
Furthermore, there are various extensions of these prior works.
For example, NEI is
more robust to the noise caused in experiments~\cite{letham2019constrained}
and SCBO is scalable to high dimensions~\cite{eriksson2021scalable}.
Another technique for constrained BO is
entropy search, such as predictive entropy search~\cite{hernandez2015predictive,garrido2023parallel}
and max-value entropy search~\cite{perrone2019constrained}.
They choose the next configuration by approximating
the expected information gain on the value of the constrained minimizer.
While entropy search could outperform c-TPE on multimodal functions
by leveraging the global search nature,
slow convergence due to the global search nature and the expensive query cost
hinder practical use.
Note that as the implementations of these methods are not provided in the aforementioned papers except for NEI, we used only NEI in the experiments.
The major advantages of TPE over standard GP-based BOs, used by all of these papers, are more natural handling of 
categorical and conditional parameters
(see Appendix~\ref{section:suppl:tpe-performance})
and easier integration of
cheap-to-evaluate partial observations
due to the linear time complexity with respect to $|\D|$.
The concept of the integration of partial observations and its results, which showed a further acceleration of c-TPE, are available in Appendix~\ref{suppl:section:knowledge-augmentation}.

Also in the evolutionary algorithm (EA) community,
constrained optimization has been studied actively, such as 
genetic algorithms (e.g. CNSGA-II~\cite{deb2002fast}),
CMA-ES~\cite{arnold20121}, or differential evolution~\cite{mezura2006modified}.
Although CMA-ES has demonstrated the best performance among
more than 100 methods for various black-box optimization problems~\cite{loshchilov2013bi}, it does not support categorical parameters, so we did not include it in our experiments.
Furthermore, since EAs have many control parameters,
such as mutation rate and population size, meta-tuning may be necessary.
Another downside of EAs is that it is hard to
integrate partial observations because
EAs require all the metrics to rank each configuration at each iteration.
In general, BO overcomes these difficulties as discussed in Appendix~\ref{suppl:section:knowledge-augmentation}.

%% file: sections/conclusions/main.tex
\section{Conclusion}
In this paper, we introduced c-TPE,
a new constrained BO method.
Although the AF of constrained BO and TPE can be naturally combined using Corollary~\ref{thm:ei-vs-pi-is-constant},
such a na\"ive extension fails in some circumstances as discussed in Section~\ref{main:section:methods}.
Based on the discussion,
we modified c-TPE so that
the formulation strictly generalizes TPE and falls back to it in settings with loose constraints.
Furthermore, we empirically demonstrated that our modifications help
to guide c-TPE toward overlaps between the top and feasible domains.
In our series of experiments on $9$ tabular benchmarks and with $27$ constraint settings,
we first showed that the performance of c-TPE is not degraded over various constraint levels
while the other BO methods we evaluated (HM2 and NEI) degraded as constraints became looser.
Furthermore, the proposed method outperformed the other methods with statistical significance;
however, since we focus only on the tabular benchmarks to enable the stability analysis
of the performance variations depending on constraint levels,
we discuss other possible situations where c-TPE might not perform well in Appendix~\ref{section:suppl:limitations}.
Since TPE is very versatile and prominently used in several active OSS tools,
such as Optuna and Ray,
c-TPE will yield a direct positive impact on practitioners in the future.
We encourage practitioners to utilize our official implementation, which is now available on OptunaHub~\cite{ozaki2025optunahub}~\footnote{
  The c-TPE package page: \url{https://hub.optuna.org/samplers/ctpe/}
}.

%% file: sections/misc/acknowledgements.tex
\section*{Acknowledgments}
The authors appreciate the valuable contributions of the anonymous reviewers and helpful feedback from Edward Bergman and Noor Awad.
Robert Bosch GmbH is acknowledged for financial support.
The authors also acknowledge funding by European Research Council (ERC) Consolidator Grant ``Deep Learning 2.0'' (grant no.\ 101045765).
Views and opinions expressed are however those of the authors only and do not necessarily reflect those of the European Union or the ERC.
Neither the European Union nor the ERC can be held responsible for them.

\begin{center}
  \includegraphics[width=0.3\textwidth]{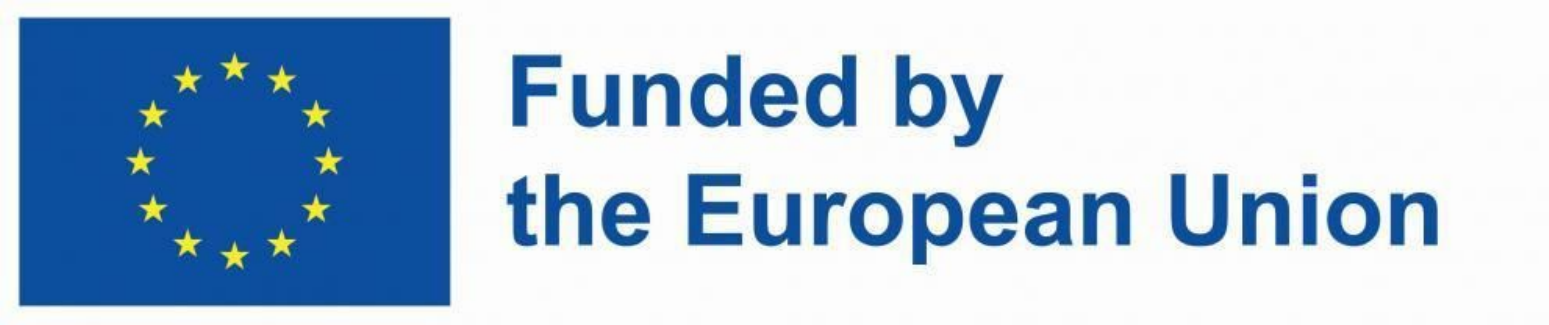}
\end{center}

%% file: settings/appendix-information.tex
\ifappendix
\clearpage
\appendix
\input{appendices/appendix-full-contents.tex}

\else

\customlabel{appendix:section:proofs}{A}
\customlabel{appx:subsection:preliminaries}{A.1}
\customlabel{appx:section:assumptions}{A.2}
\customlabel{appx:proof:section:pi-is-ei}{A.3}
\customlabel{appx:proof:eq:pi-form}{12}
\customlabel{appx:proof:eq:ei-form}{13}
\customlabel{ratio-of-ei-and-pi-is-constant}{14}
\customlabel{appx:proof:section:eci-is-product-of-relative-density-ratios}{A.4}
\customlabel{appx:proof:section:tighter-constraint-has-more-priority}{A.5}
\customlabel{appx:proofs:proof:tighter-constraint-has-more-priority}{3}
\customlabel{appx:proof:section:c-tpe-is-tpe-for-vanished-constraints}{A.6}
\customlabel{lemma-c-gamma-is-larger-than-gamma}{1}
\customlabel{all-gamma-must-be-one}{2}
\customlabel{suppl:section:split-criterion}{B}
\customlabel{fig:suppl:split-criterion}{5}
\customlabel{suppl:section:knowledge-augmentation}{C}
\customlabel{appx:hard-constraints:section}{D}
\customlabel{fig:suppl:too-greedy-for-objective-imagenet}{6}
\customlabel{suppl:tab:KA-test}{3}
\customlabel{section:suppl:limitations}{E}
\customlabel{alg:c-tpe-with-ka-algo}{2}
\customlabel{line:start-partial-observations-collection}{4}
\customlabel{line:end-partial-observations-collection}{6}
\customlabel{line:start-knowledge-augmentation}{12}
\customlabel{line:end-knowledge-augmentation}{15}
\customlabel{fig:suppl:violation-over-time}{7}
\customlabel{appx:limitations:toy-problem}{8}
\customlabel{appx:limitations:eq:toy-example}{24}
\customlabel{section:suppl:tpe-performance}{F}
\customlabel{fig:suppl:tpe-performance}{9}
\customlabel{tab:tpe-performance}{4}
\customlabel{appx:experiment-settings:section:details}{G}
\customlabel{section:suppl:robustness-to-percentile}{H}
\customlabel{section:suppl:rank-over-time}{I}
\customlabel{suppl:fig:10-50-90-eval-vs-loss-hpolib-nparams}{10}
\customlabel{suppl:fig:10-50-90-eval-vs-loss-hpolib-runtime}{11}
\customlabel{suppl:fig:10-50-90-eval-vs-loss-hpolib-nparams-runtime}{12}
\customlabel{suppl:fig:10-50-90-eval-vs-loss-nb101-nparams}{13}
\customlabel{suppl:fig:10-50-90-eval-vs-loss-nb101-runtime}{14}
\customlabel{suppl:fig:10-50-90-eval-vs-loss-nb101-nparams-runtime}{15}
\customlabel{suppl:fig:10-50-90-eval-vs-loss-nb201-nparams}{16}
\customlabel{suppl:fig:10-50-90-eval-vs-loss-nb201-runtime}{17}
\customlabel{suppl:fig:10-50-90-eval-vs-loss-nb201-nparams-runtime}{18}
\customlabel{suppl:fig:eval-vs-avgrank-nparams}{19}
\customlabel{suppl:fig:eval-vs-avgrank-runtime}{20}
\customlabel{suppl:fig:eval-vs-avgrank-nparams-runtime}{21}

\fi

%% file: appendices/appendix-full-contents.tex
\input{appendices/proofs/main.tex}

\input{appendices/algorithm-intuition/main.tex}
\input{appendices/knowledge-augmentation/main.tex}
\input{appendices/hard-constraints/main.tex}
\input{appendices/limitations/main.tex}
\input{appendices/tpe-performance/main.tex}
\input{appendices/experiments/main.tex}

\bibliographystyleappx{bib-style}  
\bibliographyappx{ref}

%% file: appendices/proofs/main.tex
\section{Proofs}
\label{appendix:section:proofs}

\input{appendices/proofs/preliminaries.tex}

\input{appendices/proofs/assumptions.tex}

\input{appendices/proofs/pi-is-ei.tex}

\input{appendices/proofs/eci-is-product-of-relative-density-ratios.tex}

\input{appendices/proofs/tighter-constraint-has-more-priority.tex}

\input{appendices/proofs/c-tpe-is-tpe-for-vanished-constraints.tex}

%% file: appendices/proofs/preliminaries.tex
\vspace{0mm}
\subsection{Preliminaries}
\label{appx:subsection:preliminaries}
\vspace{0mm}
We use the following definitions to make the discussion of constraint levels simpler:
\vspace{0mm}
\begin{definition}[$\gamma$-quantile value]
  Given a quantile $\gamma \in (0, 1]$
  and a measurable function $f\mathrm{:}~\Xv \rightarrow \mathbb{R}$,
  $\gamma$-quantile value $f^\gamma \in \mathbb{R}$
  is a real number such that$\mathrm{:}$
  \vspace{0mm}
  \begin{equation}
    \begin{aligned}
      f^\gamma \coloneqq \inf\biggl\{f^\star \in \mathbb{R} ~\biggl|
      \int_{\xv \in \Xv} \mathbbm{1}[f(\xv) \leq f^\star]
      \frac{\mu(d\xv)}{\mu(\Xv)} \geq \gamma
      \biggr\}.
    \end{aligned}
  \end{equation}
  where $\mu$ is the Lebesgue measure on $\Xv$.
\end{definition}
\vspace{0mm}
\begin{definition}
  Given a constraint $c~\!\mathrm{:}~ \Xv \rightarrow \mathbb{R}$
  and a constraint threshold $c^\star \in \mathbb{R}$,
  $\gcistar{}$ is defined as the quantile of
  the constraint $c$ such that $c^\star = c^\gcistar{}$.
\end{definition}
\vspace{0mm}
\begin{definition}[$\Gamma$-feasible domain]
  Given a set of constraint thresholds $\cistar{i} \in \mathbb{R}$ $\mathrm{(}$for $i \in \{1,\dots, C\}\mathrm{)}$,
  we define the feasible domain $\Xv^\prime = \{\xv \in \Xv | \forall i, c_i(\Xv) \leq \cistar{i} \}$.
  Then the feasible domain ratio is computed as $\Gamma = \mu(\Xv^\prime) / \mu(\Xv) \in (0, 1]$
  and the domain is said to be the $\Gamma$-feasible domain.
\end{definition}
\vspace{0mm}
Note that $\xv \in \mathbb{R}^D$ is a hyperparameter configuration,
$\Xv = \Xv_1 \times \dots \times \Xv_D \subseteq \mathbb{R}^D$
is the search space of the hyperparameter configurations,
$\Xv_d \subseteq \mathbb{R}~(\text{for }d = 1,\dots,D)$ is the domain of
the $d$-th hyperparameter,
Note that we consider two assumptions mentioned in Appendix~\ref{appx:section:assumptions}
and those assumptions allow the whole discussion to be extended to search spaces
with categorical parameters.

%% file: appendices/proofs/assumptions.tex
\vspace{0mm}
\subsection{Assumptions}
\label{appx:section:assumptions}
\vspace{0mm}
In this paper, we assume the following:
\begin{enumerate}
      \vspace{0mm}
      \item Objective $f: \Xv \rightarrow \mathbb{R}$
            and constraints $c_i: \Xv \rightarrow \mathbb{R}$
            are Lebesgue integrable and are measurable functions defined over the compact measurable subset
            $\Xv \subseteq \mathbb{R}^D$,
      \vspace{0mm}
      \item The support of PI for the objective $\prob(f \leq f^\star | \Xv, \D)$ and each constraint
            $\prob(c_i \leq \cistar{i} | \xv, \D)$ covers the whole domain $\Xv$
            for an arbitrary choice of $f^\star, \cistar{i} \in \mathbb{R}$,
      \vspace{0mm}
\end{enumerate}
where
$\D = \{(\xv_n, f_n, \cv_n)\}_{n=1}^N$ is a set of observations,
and $\cv_n = [c_{1,n}, \dots, c_{C,n}] \in \mathbb{R}^C$ is the $n$-th observation of each constraint.
The Lebesgue integrability easily holds
for TPE as TPE only considers the order of each configuration
and almost all functions are measurable unless they are constructive.
Note that we also assume a categorical parameter to be $\Xv_i = [1, K]$
as in the TPE implementation \citeappx{bergstra2011algorithms}
where $K$ is a number of categories.
As we do not require the continuity of $f$ and $c_i$
with respect to hyperparameters in our analysis,
this definition is valid
as long as the employed kernel
for categorical parameters
treats different categories to be equally similar
such as Aitchison-Aitken Kernel~\citeappx{aitchison1976multivariate}.
In this definition, $x, x^\prime \in \Xv_i$
are viewed as equivalent
as long as $\lfloor x \rfloor  = \lfloor x^\prime \rfloor$
and it leads to the random sampling of each category to be uniform and
the Lebesgue measure of $\Xv$ to be non-zero.

%% file: appendices/proofs/pi-is-ei.tex
\subsection{Proof of Proposition~\ref{thm:ei-vs-pi-is-constant}}
\label{appx:proof:section:pi-is-ei}
\begin{proof}
  Using $\mathrm{Eq.}~(\ref{eq:l-and-g-trick})$, $\mathrm{PI}$
  is computed as$\mathrm{:}$
  \begin{align}
    \prob(f \leq f^\star \mid \xv, \D) &= 
    \int_{-\infty}^{f^\star} p(f | \xv,\D)df \notag \displaybreak[4] \\
    &= \int_{-\infty}^{f^\star} \frac{p(\xv|f, \D)p(f| \D)}{p(\xv| \D)}df \notag\\
    &= \frac{p(\xv|\Dl)}{p(\xv | \D)}
    \int_{-\infty}^{f^\star} p(f | \D) df.\notag\\
  \label{appx:proof:eq:pi-form}
  \end{align}
  Notice that $\D$ is split by $f^\star$.
  $\EI$ in $\mathrm{TPE}$ is computed as$\mathrm{:}$
  \begin{equation}
  \begin{aligned}
    \EI_{f^\star}[\xv | \D] &=
    \frac{p(\xv | \Dl)}{p(\xv | \D)}
      \int_{-\infty}^{f^\star}
      (f^\star - f)p(f | \D) df.
  \end{aligned}
  \label{appx:proof:eq:ei-form}
  \end{equation}
  When we take the ratio of $\mathrm{Eqs}~(\ref{appx:proof:eq:pi-form}),(\ref{appx:proof:eq:ei-form})$, the part that depends on $\xv$ cancels out as follows$\mathrm{:}$
  \begin{equation}
  \begin{aligned} 
      \frac{\int_{-\infty}^{f^\star}
      (f^\star - f) p(f | \D) df}{
        \int_{-\infty}^{f^\star} p(f | \D) df}
     = \mathrm{const~w.r.t.}~\xv,
  \end{aligned}
  \label{ratio-of-ei-and-pi-is-constant}
  \end{equation}
  where, since we assume that the support of $\prob(f \leq f^\star | \xv, \D)$
  covers the whole domain $\Xv$,
  i.e. $\forall \xv \in \Xv, \prob(f \leq f^\star | \xv, \D) \neq 0$
  and $f$ is Lebesgue integrable,
  i.e. the expectation of $f$ exists and $\int |f| \mu(d\xv) < \infty$,
  both numerator and denominator always take a positive finite value,
  and thus the $\mathrm{LHS}$ of $\mathrm{Eq.}~(\ref{ratio-of-ei-and-pi-is-constant})$
  takes a finite positive constant value.
\end{proof}

%% file: appendices/proofs/eci-is-product-of-relative-density-ratios.tex
\subsection{Proof of Corollary~\ref{main:method:proof:eci-is-product-of-relative-density-ratios}}
\label{appx:proof:section:eci-is-product-of-relative-density-ratios}
\begin{proof}
  Under the $\mathrm{TPE}$ formulation,
  $\EI_{f^\star}(\xv | \D)$ is proportional to $\rel_0(\xv|\D)$
  and $\EI_{\cistar{i}}(\xv | \D)$ is proportional to $\rel_i(\xv|\D)$ as shown by Bergstra \textit{et al.} \shortcite{bergstra2011algorithms}.
  Furthermore, since $\EI$ and $\mathrm{PI}$ are equivalent in the $\mathrm{TPE}$ formulation from Proposition~\ref{thm:ei-vs-pi-is-constant},
  $\mathrm{ECI}$ for $\mathrm{TPE}$ satisfies the following using $\mathrm{Eq.~(\ref{main:method:eq:eci-transformation-derivation})}$:
  \begin{equation}
    \begin{aligned}
      \eci &\propto \prob(f\leq f^\star|\xv,\D) \prod_{i=1}^C \prob(c_i
      \leq \cistar{i}|\xv,\D) \\
      &\propto \prod_{i=0}^C \rel_i(\xv|\D).
    \end{aligned}
  \end{equation}
  This completes the proof.
\end{proof}

%% file: appendices/proofs/tighter-constraint-has-more-priority.tex
\subsection{Proof of Theorem~\ref{main:vanished-constraints:thm:tight-constraint-have-more-priority}}
\label{appx:proof:section:tighter-constraint-has-more-priority}
\begin{proof}
  From $\mathrm{Corollary~\ref{main:method:proof:eci-is-product-of-relative-density-ratios}}$,
  since $\eci \propto \prod_{k=0}^C \rel_{k}$ holds,
  the partial derivative of the $\mathrm{RHS}$ with respect to
  the density ratio $r_k(\xv)$ for $k \in \{0,\dots,C\}$ is
  computed as follows$\mathrm{:}$
  \begin{align}
    \pd{\eci}{r_k}                                & \propto  \pd{\rel_k}{r_k} \prod_{k^\prime \neq k} \rel_{k^\prime} \notag                    \\
                                                  & = \pd{}{r_k}
    \frac{1}{\hat{\gamma}_k + (1 - \hat{\gamma}_k)r_k^{-1}}\notag
    \prod_{k^\prime \neq k} \rel_{k^\prime}                                                                                                     \\
                                                  & =\frac{1 - \hat{\gamma}_k}{(\hat{\gamma}_k r_k + 1 - \hat{\gamma}_k)^2}
    \prod_{k^\prime \neq k} \rel_{k^\prime}                                                                                               \\
                                                  & = \frac{1 - \hat{\gamma}_k}{r_k^2}\rel_k \prod_{k^\prime=0}^C \rel_{k^\prime} \geq 0 \notag \\
    (\because \forall k^\prime \in \{0,\dots,C\}, & r_{k^\prime}, \rel_{k^\prime} > 0
    , 0 \leq 1 - \hat{\gamma}_k < 1).\notag
    \label{}
  \end{align}
  For this reason, the $\mathrm{LHS}$ takes zero if and only if $\hat{\gamma}_k = 1$.
  Using the result, the following holds with a positive constant number $\alpha$$\mathrm{:}$
    \begin{equation}
      \begin{aligned}
         & \pd{\eci}{r_i} - \pd{\eci}{r_j} \\
         & = \alpha
        \Biggl(
        \frac{1 - \hat{\gamma}_i}{r_i^2}\rel_i - \frac{1 - \hat{\gamma}_j}{r_j^2}\rel_j
        \Biggr)                            \\
         & = \alpha
        \Biggl(
        \frac{1 - \hat{\gamma}_i}{
            \hat{\gamma}_i r_i^2 + (1 - \hat{\gamma}_i)r_i
          } - \frac{1 - \hat{\gamma}_j}{
            \hat{\gamma}_j r_j^2 + (1 - \hat{\gamma}_j)r_j
          }
        \Biggr)                            \\
         & =  \alpha
        \Biggl(
        \biggl(
          r_i + \frac{\hat{\gamma}_i}{1 - \hat{\gamma}_i}r_i^2
          \biggr)^{-1}
        - \biggl(
          r_j + \frac{\hat{\gamma}_j}{1 - \hat{\gamma}_j}r_j^2
          \biggr)^{-1}
        \Biggr).                           \\
      \end{aligned}
    \end{equation}
    Since $r_i, r_j > 0$ holds and
    \begin{equation}
      \begin{aligned}
        r_i + \frac{\hat{\gamma}_i}{1 - \hat{\gamma}_i}r_i^2
        \geq
        r_j + \frac{\hat{\gamma}_j}{1 - \hat{\gamma}_j}r_j^2,
      \end{aligned}
    \end{equation}
    \begin{equation}
      \begin{aligned}
         & \pd{\eci}{r_i} - \pd{\eci}{r_j} \\ &= \alpha
        \Biggl(
        \biggl(
          r_i + \frac{\hat{\gamma}_i}{1 - \hat{\gamma}_i}r_i^2
          \biggr)^{-1}
        - \biggl(
          r_j + \frac{\hat{\gamma}_j}{1 - \hat{\gamma}_j}r_j^2
          \biggr)^{-1}
        \Biggr) \geq 0
      \end{aligned}
    \end{equation}
    holds.
    When we assume $\hat{\gamma}_i = \hat{\gamma}_j$
    and $r_i = r_j$,
  we get the equality.
  This completes the proof.
  \label{appx:proofs:proof:tighter-constraint-has-more-priority}
\end{proof}
Note that since $x/(1-x)$ is a monotonically increasing function
in $x \in [0, 1)$ and $\hat{\gamma}_i \leq \hat{\gamma}_j$ from the assumption,
\begin{equation}
  \begin{aligned}
    0 \leq \alpha_i \coloneqq \frac{\hat{\gamma}_i}{1 - \hat{\gamma}_i} \leq
    \alpha_j \coloneqq \frac{\hat{\gamma}_j}{1 - \hat{\gamma}_j}
  \end{aligned}
\end{equation}
holds.
Furthermore, using $r_i, r_j > 0$,
if we assume $r_i < r_j$, then $r_i < r_j, r_i^2 < r_j^2, \alpha_i < \alpha_j$
and it leads to a larger value of partial derivative in the $i$-th constraint;
therefore, $r_j$ must be smaller than $r_i$ for
its contribution to be larger than that from $r_i$.
It implies that we will not put more priority on the constraints
with large feasible domains
unless
those constraints are likely to be violated,
which means the density ratios for those constraints are small.

%% file: appendices/proofs/c-tpe-is-tpe-for-vanished-constraints.tex
\subsection{Proof of Corollary~\ref{thm:ctpe-converges-to-single-objective}}
\label{appx:proof:section:c-tpe-is-tpe-for-vanished-constraints}

To prove Corollary~\ref{thm:ctpe-converges-to-single-objective}, we first show two lemmas.
\begin{lemma}
  Given a $\Gamma$-feasible domain ($\Gamma > 0$) with
  constraint thresholds of $\cistar{i}$ for all $i \in \{1, \dots, C\}$,
  each constraint satisfies
  \begin{equation}
    \forall i \in \{1, \dots, C\}, \gamma_i \geq \Gamma.
  \end{equation}
  \label{lemma-c-gamma-is-larger-than-gamma}
\end{lemma}
\vspace{-4mm}
\begin{proof}
  Let the feasible domain for the $i$-th constraint be
  $\Xv^\prime_i = \{\xv \in \Xv | c_i \leq \cistar{i}\}$.
  Then the feasible domain is $\Xv^\prime = \bigcap_{i=1}^C \Xv^\prime_i$.
  Since $\Xv^\prime_i$
  is a measurable set by definition
  and $\Xv^\prime \subseteq \Xv_i^\prime$ holds,
  $\Gamma / \gamma_i = \mu(\Xv^\prime) / \mu(\Xv^\prime_i) \leq 1$
  holds.
  $\Gamma$ is a positive number, so $\gamma_i \geq \Gamma$
  and this completes the proof.
\end{proof}
\begin{lemma}
  The domain is $\mathrm{(}\Gamma = 1\mathrm{)}$-feasible domain iff:
  \begin{equation}
    \forall i \in \{1, \dots, C\}, \gamma_i = 1.
  \end{equation}
  \label{all-gamma-must-be-one}
\end{lemma}
\vspace{-4mm}
\begin{proof}
  Suppose $\gamma_i < 1$ for some $i \in \{1, \dots, C\}$,
  since we immediately obtain $\Gamma \leq \gamma_i < 1$
  from $\mathrm{Lemma~\ref{lemma-c-gamma-is-larger-than-gamma}}$,
  the assumption does not hold.
  For this reason, $\gamma_i \geq 1$
  for all $i \in \{1, \dots, C\}$
  and since $\gamma_i \leq 1$ by definition,
  $\gamma_i = 1$
  for all $i \in \{1, \dots, C\}$.
  Since $\Xv_i^\prime = \Xv$ for all $i \in \{1,\dots,C\}$, $\bigcup_{i=1}^C \Xv_i^\prime = \bigcup_{i=1}^C \Xv = \Xv$ holds.
  This completes the proof.
\end{proof}

Using Lemma~\ref{all-gamma-must-be-one} and Theorem~\ref{main:vanished-constraints:thm:tight-constraint-have-more-priority},
we prove Corollary~\ref{thm:ctpe-converges-to-single-objective}.
\begin{proof}
  From $\mathrm{Lemma~\ref{all-gamma-must-be-one}}$,
  when $\Gamma = 1$ holds,
  $\gamma_i = 1$ for all $i \in \{1, \dots, C\}$
  $\mathrm{}$ holds,
  and we plug $\gamma_i = 1$ into $\mathrm{Theorem~\ref{main:vanished-constraints:thm:tight-constraint-have-more-priority}}$.
  Then we obtain$\mathrm{:}$
  \begin{equation}
  \begin{aligned}
    \forall i \in \{1, \dots, C\}, \pd{\eci}{r_i} = 0.
  \end{aligned}
  \end{equation}
  For this reason,
  $\eci \propto \prod_{i=0}^C \rel_i(\xv|\D) \propto \rel_0(\xv|\D) \rank r_0(\xv|\D)$
  and this completes the proof.
\end{proof}

%% file: appendices/algorithm-intuition/main.tex
\begin{figure*}[t]
  \centering
  \includegraphics[width=0.98\textwidth]{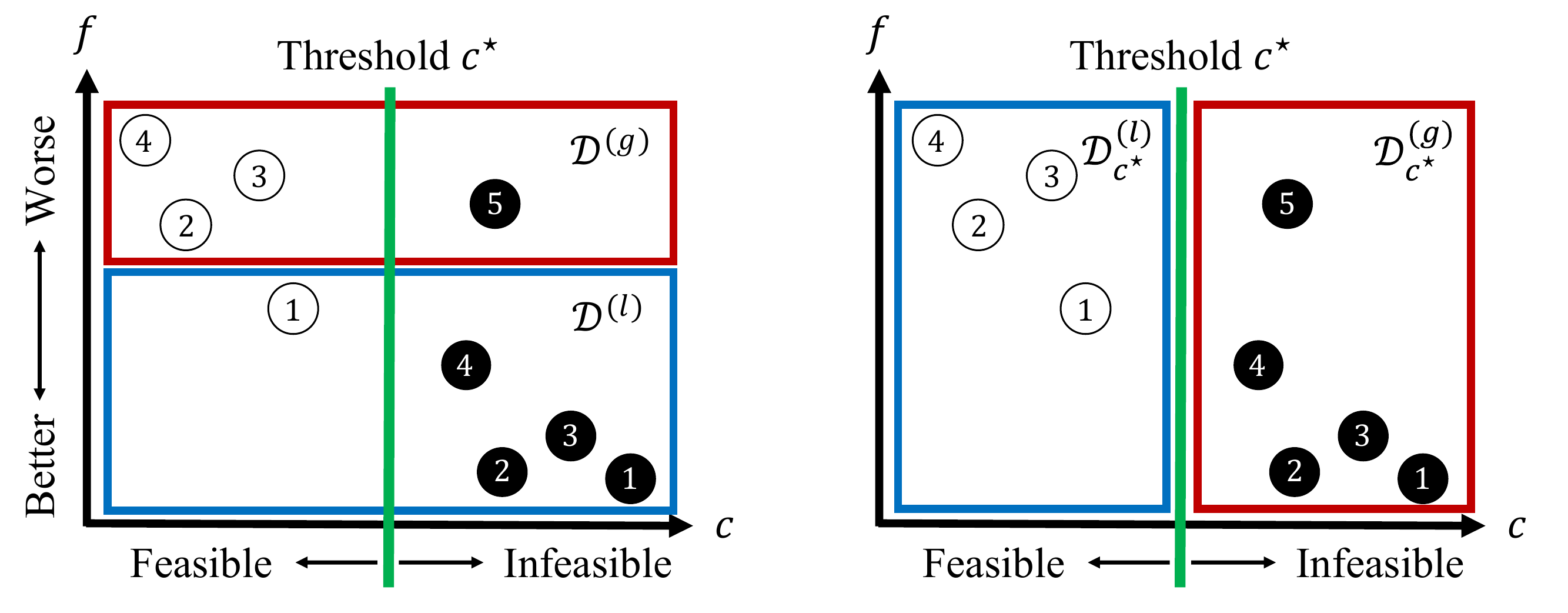}
  \vspace{-2mm}
  \caption{The conceptual visualizations of
  the split algorithm for the objective and
  for each constraint.
  The black and white circles represent
  infeasible and feasible solutions, respectively.
  The numberings for white and black circles
  stand for the ranking of the objective value among
  feasible and infeasible solutions, respectively.
  The configurations enclosed by the red and blue rectangles
  belong to the \emph{bad} and \emph{good} groups, respectively.
  \textbf{Left}: the split for the objective.
  While the original algorithm is supposed to take only
  the black circle (infeasible solution) with 1 ($= \lceil \sqrt{9}/4 \rceil$) for the good group,
  our algorithm takes until the white circle (feasible solution)
  with 1, and thus we include the black circles till 4 as well.
  \textbf{Right}: the split for the constraint.
  Our algorithm takes all white circles (feasible solutions).
  }
  \label{fig:suppl:split-criterion}
  \vspace{-2mm}
\end{figure*}

\begin{figure*}[t]
  \centering
  \includegraphics[width=0.96\textwidth]{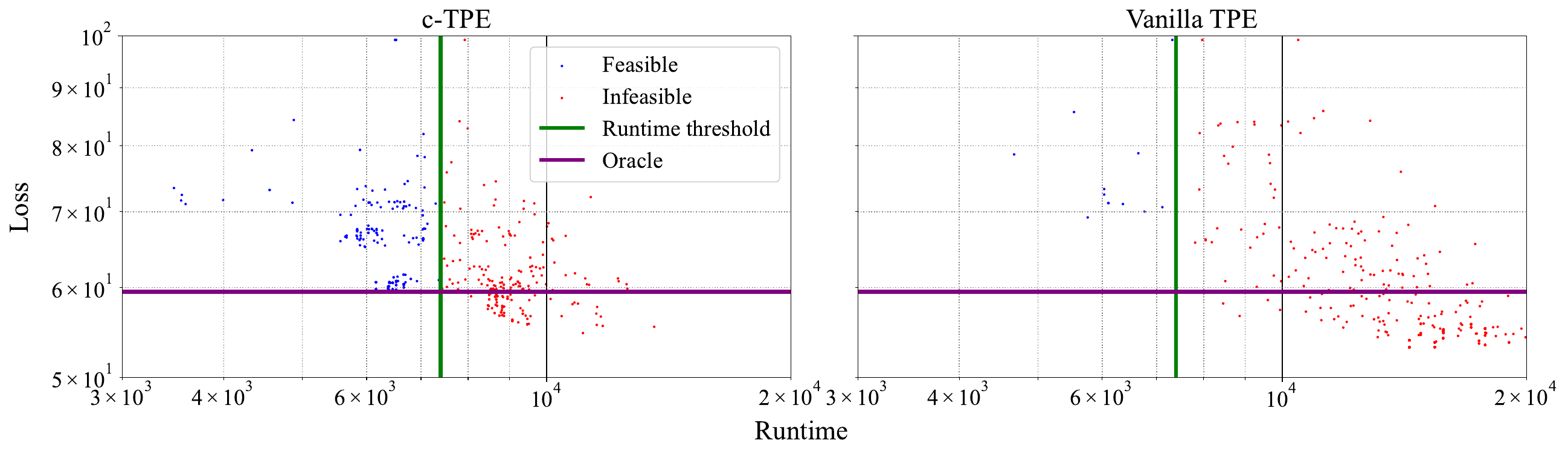}
  \vspace{-3mm}
  \caption{
    The visualization of the observations obtained by
    c-TPE and the vanilla TPE on ImageNet16-120 of NAS-Bench-201 with $\truegamma = 0.1$.
    Each optimization was run $5$ times.
    The red and blue dots are infeasible and feasible solutions, respectively.
    The left side of the threshold (green line) is the feasible domain.
    The goal is to find configurations near the oracle (purple line) in the feasible domain.
    \textbf{Left}: the optimization by c-TPE.
    Red dots (infeasible solutions) locate relatively close to
    the threshold (green line). 
    \textbf{Right}: the optimization by the vanilla TPE.
    Since red dots (infeasible solutions) locate far away from
    the threshold (green line),
    it cannot intensively search near the threshold.
  }
  \vspace{-3mm}
  \label{fig:suppl:too-greedy-for-objective-imagenet}
\end{figure*}

\section{Further Details of Split Algorithms}
\label{suppl:section:split-criterion}
In this section, we describe the intuition and more details on how
the split algorithm works.

\subsection{Split Algorithm of Objective}
Figure~\ref{fig:suppl:split-criterion} presents
how to split observations into good and bad groups.
For the split for the objective (\textbf{Left}),
as there are $N = 9$ observations, 
we will include
$\lceil \sqrt{N} / 4 \rceil = \lceil \sqrt{9} / 4 \rceil = 1$ feasible solution in
$\Dl$.
We first find the feasible observation
with the best objective value, which is the white-circled observation 1.
Then $\Dl_0$ and $\Dg_0$ are obtained by splitting the observations at the white observation 1
along the horizontal axis.
As discussed in Section~\ref{main:pitfall-issue02:section:issue02},
this modification is effective for small overlaps
and small overlaps are caused by observations with the best objective values far away from the feasible domain,
e.g. the black-circled observations 1 and 3 in Figure~\ref{fig:suppl:split-criterion}.
For example, Figure~\ref{fig:suppl:too-greedy-for-objective-imagenet}
visualizes the observations by c-TPE and the vanilla TPE on
ImageNet16-120 of NAS-Bench-201 with $\truegamma = 0.1$.
There are many observations with strong performance than the oracle (purple line)
that are far from the feasible domain (left side of the green line)
in the result of the vanilla TPE.
Note that the oracle is the best objective value that satisfies the constraint.
Theorem~\ref{main:vanished-constraints:thm:tight-constraint-have-more-priority} guarantees that c-TPE will not prioritizes such observations.

\subsection{Split Algorithm of Each Constraint}
We first note that we show, for simplicity, a 1D example
and abbreviate $c_i, \cistar{i}, \Dl_i, \Dg_i$
as $c, c^\star, \Dl_{c^\star}, \Dg_{c^\star}$, respectively.
For the split of each constraint (\textbf{Right}),
we take the observations
with constraint values less than $c^\star$ into
$\Dl_{c^\star}$ (inside the blue rectangle) and vice versa.
When there are no observations in the feasible domain,
we only take the observation with the best constraint value
among all the observations into $\Dl_{c^\star}$ and
the rest into $\Dg_{c^\star}$.
Since this selection increases the priority of this constraint
as mentioned in Theorem~\ref{main:vanished-constraints:thm:tight-constraint-have-more-priority},
it raises the probability of yielding feasible solutions quickly.

%% file: appendices/knowledge-augmentation/main.tex
\begin{algorithm}[t]
  \caption{c-TPE with knowledge augmentation}
  \label{alg:c-tpe-with-ka-algo}
  \begin{algorithmic}[1]
    \State{$N_{\mathrm{init}}$ (The number of initial configurations), $N_s$ (The number of candidates to consider in the optimization of the AF), $N_p$ (The number of configurations for KA)}
    \Comment{Control parameters}
    \State{$I = \{i_j\}_{j=1}^{C_p}$}
    \Comment{Indices of cheap constraints}
    \State{$\D_{p} \leftarrow \emptyset, \D \leftarrow \emptyset$}
    \For{$n = 1, \dots, N_p$} \Comment{Collect cheap information}
    \label{line:start-partial-observations-collection}
    \State{Randomly pick $\xv$}
    \State{\textcolor{cyan}{$\D_p \leftarrow \D_p \cup \{(\xv, f(\xv), c_{i_1}(\xv), \dots, c_{i_{C_p}}(\xv))\}$}}
    \EndFor
    \label{line:end-partial-observations-collection}
    \For{$n = 1, \dots, N_{\mathrm{init}}$}
    \State{Randomly pick $\xv$}
    \State{$\D \leftarrow \D \cup \{(\xv, f(\xv), c_1(\xv), \dots, c_C(\xv))\}$}
    \EndFor
    \While{Budget is left}
    \For{$i = 0, \dots, C$}
    \If{$i \in I$}
    \label{line:start-knowledge-augmentation}
    \State{\textcolor{cyan}{$\D_{\text{aug}} = \D \cup \D_{p}$}}
    \Else
    \State{$\D_{\text{aug}} = \D$}
    \EndIf
    \label{line:end-knowledge-augmentation}
    \State{Split $\D_{\text{aug}}$ into $\Dl_i$ and $\Dg_i$, $\hat{\gamma}_i \leftarrow |\Dl_i| / |D_{\text{aug}}|$}
    \State{Build $p(\cdot | \Dl_i), p(\cdot | \Dg_i)$}
    \State{$\{\xv_j\}_{j=1}^{N_s} \sim p(\cdot | \Dl_i), \mathcal{S} \leftarrow \mathcal{S} \cup \{\xv_j\}_{j=1}^{N_s}$}
    \EndFor
    \State{Pick $\xopt \in \argmax_{\xv \in \mathcal{S}} \prod_{i=0}^C \rel_i(\xv|\D)$}
    \State{$\D \leftarrow \D \cup \{(\xopt, f(\xopt), c_1(\xopt), \dots, c_C(\xopt))\}$}
    \EndWhile
  \end{algorithmic}
\end{algorithm}

\begin{figure}[b]
  \centering
  \includegraphics[width=0.48\textwidth]{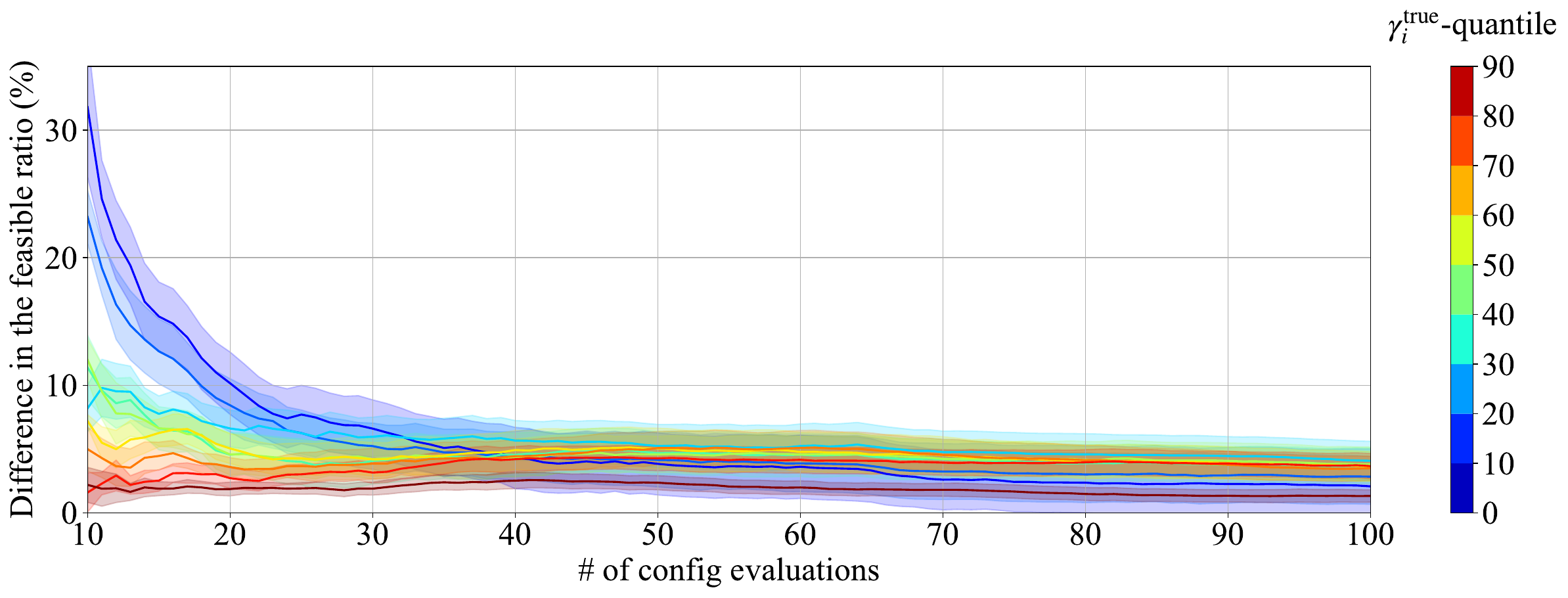}
  \vspace{-4mm}
  \caption{The effect of KA in the optimizations with a constraint for network size.
  The horizontal axis shows the number of evaluated configurations in optimizations
  and the vertical axis shows the difference in the cumulated ratio of feasible solutions,
  i.e. the ratio of the number of feasible solutions to that of the whole observations,
  between c-TPE and c-TPE with KA using 200 randomly sampled configurations.
  The weak-color bands show the standard error of
  mean values of 50 runs for 9 benchmarks.
  }
  \vspace{-2mm}
  \label{fig:suppl:violation-over-time}
\end{figure}

\section{Integration of Partial Observations}
\label{suppl:section:knowledge-augmentation}
In this section, we discuss the integration of partial observations
for BO and we name the integration ``Knowledge augmentation (KA)''.

\subsection{Knowledge Augmentation}
When some constraints can be precisely evaluated
with a negligible amount of time compared to others,
practitioners typically would like to use KA.
For example, the network size of deep learning models is trivially
computed in seconds while the final validation performance requires
several hours to days.
In this case, we can obtain many observations only for network size
and augment the knowledge of network size prior to the optimization
so that the constraint violations will be reduced in the early stage
of optimizations.

To validate the effect of KA,
all of the additional results in the appendix include
the results obtained using c-TPE with KA.
In the experiments, we augmented the knowledge only for network size
and we did not include runtime as a target of KA
because although runtime can be roughly estimated from a 1-epoch training,
such estimations are not precise.
However, practitioners can include such rough estimations into
partial observations as long as
they can accept errors caused by them.

\subsection{Algorithm of Knowledge Augmentation}
Algorithm~\ref{alg:c-tpe-with-ka-algo} is
the pseudocode of c-TPE with KA.
We first need to specify a set of indices for cheap constraints
$I = \{i_j\}_{j=1}^{C_p}$ where $C_p (< C)$ is
the number of cheap constraints
and $I \subseteq \{1, \dots, C\}$.
In Lines~\ref{line:start-partial-observations-collection} -- \ref{line:end-partial-observations-collection},
we first collect partial observations $\D_{p}$.
Then we augment observations
in Lines~\ref{line:start-knowledge-augmentation} -- \ref{line:end-knowledge-augmentation}
if partial observations are available for the corresponding constraint.
We denote the augmented set of observations $\D_{\mathrm{aug}}$.
When the AF follows
Eq.~(\ref{eq:expected-constraint-simple}),
the predictive models for each constraint are independently trained
due to conditional independence.
It enables us to introduce different amounts of observations for each constraint.
Since c-TPE follows Eq.~(\ref{eq:expected-constraint-simple}),
we can employ KA.
As discussed in Section~\ref{main:section:related-works},
it is hard to apply KA to evolutionary algorithms due to their algorithm nature
and KA causes a non-negligible bottleneck for GP-based BO as the number of observations grows.

\subsection{\normalsize{Empirical Results of Knowledge Augmentation}}

In this experiment, we optimized each benchmark with a constraint for network size,
and constraints for runtime and network size.
To see the effect, we measured how much KA
increases the chance of drawing feasible solutions
and tested the performance difference by the Wilcoxon signed-rank test
on 18 settings ($9$ benchmarks $\times$ $2$ constraint choices).
According to Figure~\ref{fig:suppl:violation-over-time},
the tighter the constraint becomes, the more KA helps
to obtain feasible solutions, especially in the early stage of the optimizations.
Additionally,
Table~\ref{suppl:tab:KA-test} shows
the statistically significant speedup effects of KA in $\truegamma = 0.1$.
Although KA did not exhibit the significant speedup in loose constraint levels,
it did not deteriorate the optimization quality significantly.
At the later stage of the optimizations, the effect gradually decays as
c-TPE becomes competent enough to detect violations.
In summary, KA significantly accelerates optimizations
with tight constraints and it does not 
deteriorate the optimization quality in general,
so it is practically recommended to use KA as much as possible.

\input{appendices/knowledge-augmentation/test-knowledge-augmentation.tex}

%% file: appendices/knowledge-augmentation/test-knowledge-augmentation.tex
\addtolength{\tabcolsep}{-4pt}
\begin{table*}[t]
  \begin{center}
    \caption{
      The table shows (Wins/Loses/Ties) of c-TPE with KA against c-TPE
      for optimizations with different constraint levels.
      Non-bold numbers indicate $p < 0.05$ of
      the hypothesis ``c-TPE is better than c-TPE with KA''
      by the Wilcoxon signed-rank test.
    }
    \vspace{1mm}
    \label{suppl:tab:KA-test}
    \begin{tabular}{lcccc|cccc|cccc}
      \toprule
      \multicolumn{1}{c}{Quantiles}
                      & \multicolumn{4}{c}{$\truegamma = 0.1$}
                      & \multicolumn{4}{c}{$\truegamma = 0.5$}
                      & \multicolumn{4}{c}{$\truegamma = 0.9$}
      \\
      \multicolumn{1}{c}{\# of configs}
                      & 50                                     & 100    & 150            & 200            & 50              & 100             & 150            & 200            & 50             & 100             & 150            & 200
      \\
      \midrule
      Wins/Loses/Ties & 12/5/1                                 & 11/5/2 & \textbf{7/5/6} & \textbf{6/6/6} & \textbf{6/12/0} & \textbf{5/11/2} & \textbf{7/6/5} & \textbf{5/5/8} & \textbf{9/9/0} & \textbf{10/6/2} & \textbf{8/5/5} & \textbf{6/9/3} \\
      \bottomrule
    \end{tabular}
  \end{center}
  \vspace{-3mm}
\end{table*}
\addtolength{\tabcolsep}{4pt}

%% file: appendices/hard-constraints/main.tex
\section{Hard-Constrained Optimization Problems}
\label{appx:hard-constraints:section}
In this paper, although we only handled optimization problems with inequality constraints, c-TPE is applicable to optimization problems with a hard constraint, which practitioners often face in practice.
For example, we are able to perform the training of a machine learning model with a hyperparameter configuration $\xv$ only if the memory requirement is lower than the RAM capacity of the system.
In this case, when the training with the hyperparameter configuration $\xv$ fails, we only know that the hyperparameter configuration $\xv$ does not satisfy the constraint and we do not yield either $f(\xv)$ or $c(\xv)$.
Since we only need to be able to split observations into $\Dl_i$ and $\Dg_i$ with respect to the hard constraint, $\Dl_i$ will collect all observations that satisfy the hard constraint and $\Dg_i$ will collect the others.
For the objective $f(\xv)$, we simply ignore all the observations that violate the hard constraint.
As discussed in Appendix~\ref{suppl:section:knowledge-augmentation}, the surrogate model could be simply obtained even from a set of partial observations.
In this problem setting, the feasible observations for the hard constraint $\Dl_i$ could often be an empty set and then $p(\xv | \Dl_i)$ becomes simply the non-informative prior employed in TPE~\citeappx{watanabe2023tpe}.
As $p(\cdot | \Dg_i)$ still provides the information about the violation of the hard constraint, c-TPE simply searches the regions far from the current violated observations.

%% file: appendices/limitations/main.tex
\begin{figure}
      \centering
      \includegraphics[width=0.49\textwidth]{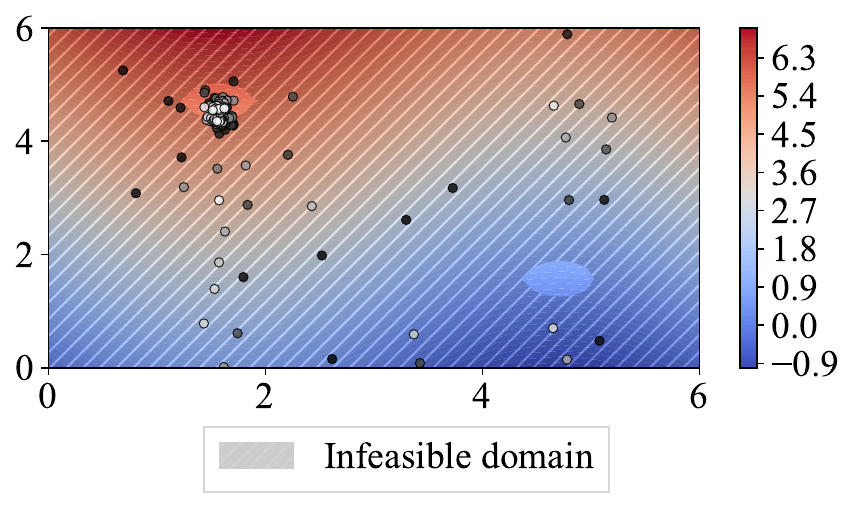}
      \caption{
            The optimization of Eq.~(\ref{appx:limitations:eq:toy-example}) by c-TPE.
            Blue implies that the objective function is better and the shaded area is infeasible domain.
            Earlier observations are colored black and later observations are colored white.
            This figure has $100$ observations.
            Since this problem has two modals and each of them is very small, c-TPE was trapped by one of the two modals and could not escape from there within the specified number of observations.
      }
      \label{appx:limitations:toy-problem}
\end{figure}

\section{Limitations}
\label{section:suppl:limitations}
In this paper, we focused on tabular benchmarks
for search spaces with categorical parameters
and with one or two constraints.
We chose the tabular benchmarks to enable the stability analysis
of the performance variations depending on constraint levels.
Furthermore, such settings are common in HPO of deep learning.
However, since practitioners may use c-TPE for other settings, we would like to discuss the following settings which
we did not cover in the paper:
\begin{enumerate}
  \item \textbf{Extremely small feasible domain size},
        \vspace{-1mm}
  \item \textbf{Many constraints},
        \vspace{-1mm}
  \item \textbf{Parallel computation}, and
        \vspace{-1mm}
  \item \textbf{Synthetic functions}.
\end{enumerate}

The first setting is an extremely small feasible domain size.
For example, when we have $\Gamma = 10^{-4}$ for $200$ evaluations
and use random search,
we will not get any feasible solutions with the probability of
$(1 - 10^{-4})^{200} = 0.9802\dots \simeq 98.0\%$.
Such settings are generally hard for most optimizers to find even one feasible solution.

The second setting is tasks with many constraints.
In our experiments, we have the constraints of runtime and network size.
On the other hand, there might be more constraints in other purposes.
Many constraints make the optimization harder
because the feasible domain size becomes smaller
as the number of constraints increases
due to the curse of dimensionality.
More formally,
when we define the feasible domain for the $i$-th constraint as
$\Xv_i^\prime = \{\xv \in \Xv | c_i(\xv) \leq \cistar{i}\}$,
the feasible domain size shrinks exponentially
unless some feasible domains are identical, i.e. $\Xv_i^\prime = \Xv_j^\prime$ for
some pairs $(i, j) \in \{1, \dots, C\} \times \{1, \dots, C\}$ such that $i \neq j$.
This setting is also generally hard due to the small feasible domain size.

The third setting is parallel computation.
In HPO, since objective functions are usually expensive,
it is often preferred to be able to optimize in parallel with less regret.
For example, since evolutionary algorithms evaluate a fixed number $G$ of configurations in one generation,
they optimize the objective function without any loss compared to the sequential setting
up to $G$ parallel processes.
Although TPE (and c-TPE) are applicable to asynchronous settings,
we cannot conclude c-TPE works nicely in parallel settings from our experiments.

The fourth setting is synthetic function.
We did not handle synthetic function
because it is hard to prepare the exact $\truegamma$.
As mentioned earlier, one of the most important points of our method
is the robustness with respect to various constraint levels.
As synthetic functions are designed to be hard in certain constraint thresholds,
it was hard to maintain the difficulties for different $\truegamma$
and to even analytically compute $\truegamma$.
It is worth noting that c-TPE is likely to not perform well on multi-modal functions.
For example, Figure~\ref{appx:limitations:toy-problem} presents such an instance.
This example uses:
\begin{equation}
\begin{aligned}
      \min_{(x_1, x_2) \in [0, 2\pi] \times [0, 2\pi]} &\sin x_1 + x_2 \\
      \mathrm{subject~to~} \sin x_1 &\sin x_2 \leq -0.95.
\end{aligned}
\label{appx:limitations:eq:toy-example}
\end{equation}
In this case, c-TPE was trapped in one of the two feasible domain where we have worse objective values.
Since this case has small feasible domains and c-TPE searches locally due to the nature of PI, it intensively searches one of the feasible domains which c-TPE first finds and it is hard for c-TPE to find both of the two modals.
In this example, c-TPE may require more evaluations to cover both modals compared to global search methods although this issue could be addressed by multiple runs of c-TPE.

Since we did not test c-TPE on those settings,
practitioners are encouraged to compare c-TPE with other methods
if their tasks of interest have the characteristics described above.

%% file: appendices/tpe-performance/main.tex
\section{Performance of Vanilla TPE}
\label{section:suppl:tpe-performance}
As described in Appendix~\ref{appx:experiment-settings:section:details},
since our TPE implementation uses multivariate kernel density estimation,
it is different from the Hyperopt implementation
that is used in most prior works.
For this reason, we compare our the performance of our TPE implementation
with that of Hyperopt, and other BO methods.
Since all settings include categorical parameters, we compare the following BO methods which are known to perform well on search space with categorical parameters.
\begin{enumerate}
  \item \textbf{TuRBO}~\citeappx{eriksson2019scalable}~\footnote{Implementation: \url{https://github.com/uber-research/TuRBO}}, and
  \item \textbf{CoCaBO}~\citeappx{ru2020bayesian}~\footnote{Implementation: \url{https://github.com/rubinxin/CoCaBO_code}}.
\end{enumerate}
CoCaBO is a BO method that focuses on the handling of categorical parameters
and TuRBO is one of the strongest BO methods developed recently.
Both methods follow the default settings.
Note that as both methods
are either not extended to constrained optimization
or not publicly available, we could not include those methods in Section~\ref{main:section:experiments}.

\begin{figure}[t]
  \centering
  \includegraphics[width=0.49\textwidth]{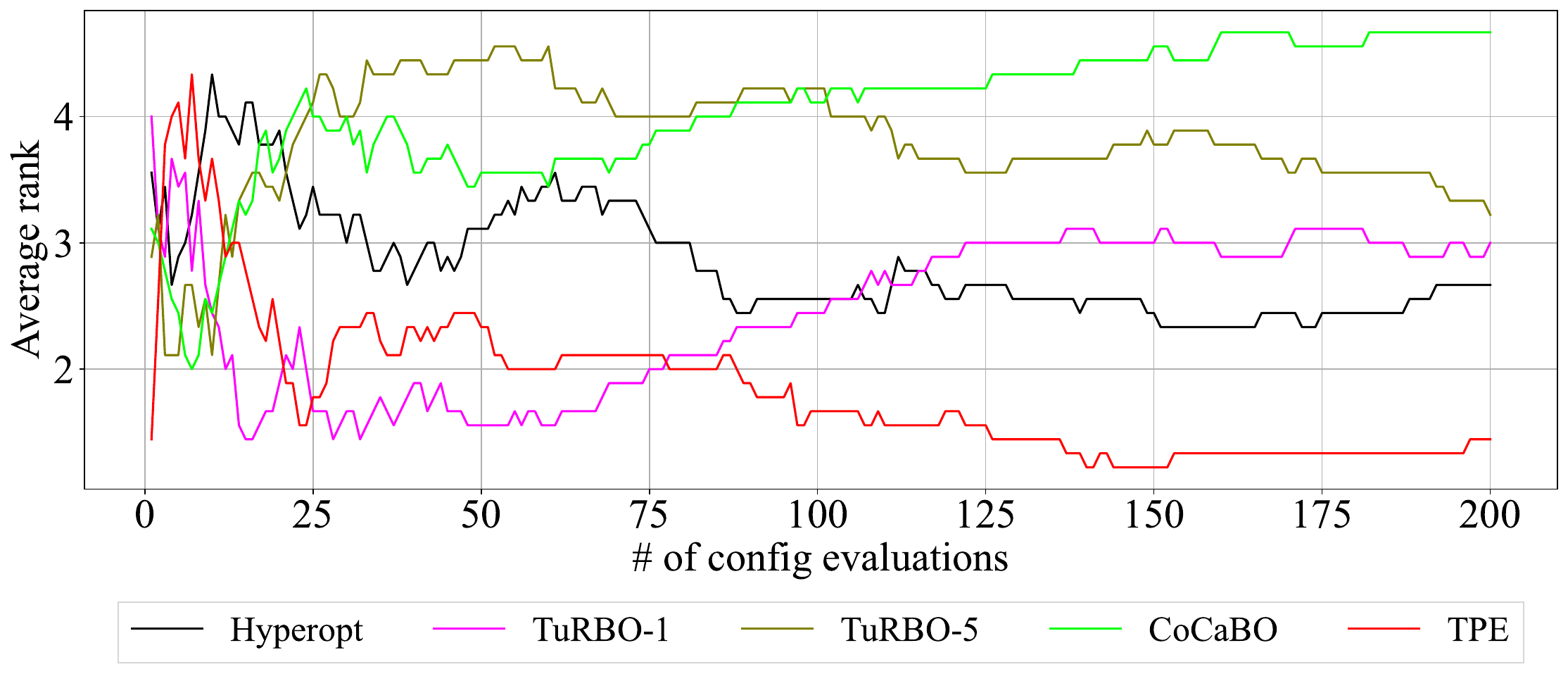}
  \caption{
    The performance comparison of our TPE implementation
    against prior works and Hyperopt implementation.
    The horizontal axis represents the number of configurations
    and the vertical axis represents the average rank of
    each method over $9$ benchmarks that were used in
    Section~\ref{main:section:experiments}.
  }
  \label{fig:suppl:tpe-performance}
\end{figure}

Figure~\ref{fig:suppl:tpe-performance} shows the average rank
over time for each method.
As seen in the figure, our TPE outperformed Hyperopt.
Furthermore, while our TPE is significantly
better than other methods in most settings,
Hyperopt is better than only CoCaBO.
On the other hand, TuRBO-1 performs better
in the early stage of optimizations although
our TPE outperforms TuRBO-1 with statistical significance,
and this cold start in the vanilla TPE might be a trade-off.
Notice that since most BO papers test performance on toy functions
and we use the tabular benchmarks,
the discussion here does not generalize and
the results only validate why we should use our TPE in our paper.

\begin{table}[t]
  \caption{
    In the table,
    we show the test results of
    The hypothesis ``The other method is better than our TPE''
    for the ``v.s. our TPE'' column
    and 
    the hypothesis ``The other method is better than Hyperopt''
    for the ``v.s. Hyperopt'' column
    by the Wilcoxon signed-rank test.
    For example, the ``TuRBO-1'' row in
    the ``v.s. our TPE'' column
    says ``N/N/W/W''.
    It means while we cannot draw any conclusion
    about the performance difference
    with $50, 100$ evaluations,
    TuRBO-1 is significantly worse than our TPE
    with $150, 200$ evaluations in our settings.
    Note that we chose $p < 0.01$ as the threshold.
    Each method was run over 15 random seeds.
  }
  \vspace{-1mm}
  \label{tab:tpe-performance}
  \makebox[0.49 \textwidth][c]{       
  \begin{tabular}{lc|c}
    \toprule
    Methods & v.s. our TPE & v.s. Hyperopt \\
    \# of configs & 50/100/150/200 & 50/100/150/200 \\
    \midrule
    our TPE  & --/--/--/-- & N/B/B/B \\
    Hyperopt & N/\textbf{W}/\textbf{W}/\textbf{W} & --/--/--/-- \\
    TuRBO-1  & N/N/\textbf{W}/\textbf{W} & B/N/N/N \\
    TuRBO-5  & \textbf{W}/\textbf{W}/\textbf{W}/\textbf{W} & \textbf{W}/N/N/N \\
    CoCaBO   & \textbf{W}/\textbf{W}/\textbf{W}/\textbf{W} & N/\textbf{W}/\textbf{W}/\textbf{W} \\
    \bottomrule
  \end{tabular}
  }
\end{table}

%% file: appendices/experiments/main.tex
\input{appendices/experiments/details-of-settings.tex}

\input{appendices/experiments/additional-results-for-percentile-robust.tex}

\input{appendices/experiments/additional-results-for-rank-over-time.tex}

%% file: appendices/experiments/details-of-settings.tex
\section{Details of Experiment Setup}
\label{appx:experiment-settings:section:details}
For all the methods using TPE, we used $N_s = 24$
and $N_{\mathrm{init}} = 10$,
which we obtain from the ratio ($5\%$) of the initial sample size and the number of evaluations,
as in the original paper~\citeappx{bergstra2013making}.
Furthermore, we employed the multivariate kernel and its bandwidth selection
used by the prior work~\citeappx{falkner2018bohb}.
Due to this modification,
our vanilla TPE implementation performs significantly
better than Hyperopt~\citeappx{bergstra2013making}~\footnote{Implementation:
\url{https://github.com/hyperopt/hyperopt}
} on our experiment settings,
and thus we would like to stress that
our TPE may produce better results compared to what we can expect
from prior works~\citeappx{daxberger2019mixed,deshwal2021bayesian,eggensperger2013towards,ru2020bayesian,turner2021bayesian}.
For more details,
see Appendix~\ref{section:suppl:tpe-performance}.
CNSGA-II is a genetic algorithm based constrained optimization method,
NEI is a GP-based constrained BO method with EI for noisy observations,
and HM2 is a random-forest-based constrained BO method with ECI,
which implements major parts of SMAC~\citeappx{lindauer2021smac3} to perform constrained optimization.
Note that NSGA-II has a constrained version
and we used the constrained version named CNSGA-II.
The vanilla TPE is evaluated in order to demonstrate the improvement of c-TPE
from TPE for non-constrained settings.
CNSGA-II, NEI, and HM2 followed the default settings in each package.

%% file: appendices/experiments/additional-results-for-percentile-robust.tex
\section{Additional Results for Section~\ref{section:main:experiments:robustness-to-constraint-percentile}}
\label{section:suppl:robustness-to-percentile}
In this section, we present the additional results for Section~\ref{section:main:experiments:robustness-to-constraint-percentile}
to show how robust c-TPE is over various constraint levels.
Note that we picked only network size as a cheap constraint
and did not pick runtime as discussed in Appendix~\ref{suppl:section:knowledge-augmentation},
and we used $N_p = 200$ throughout all the experiments.

\subsection{Results on HPOlib}

\begin{figure*}[p]
  \centering
  \includegraphics[width=0.7\textwidth]{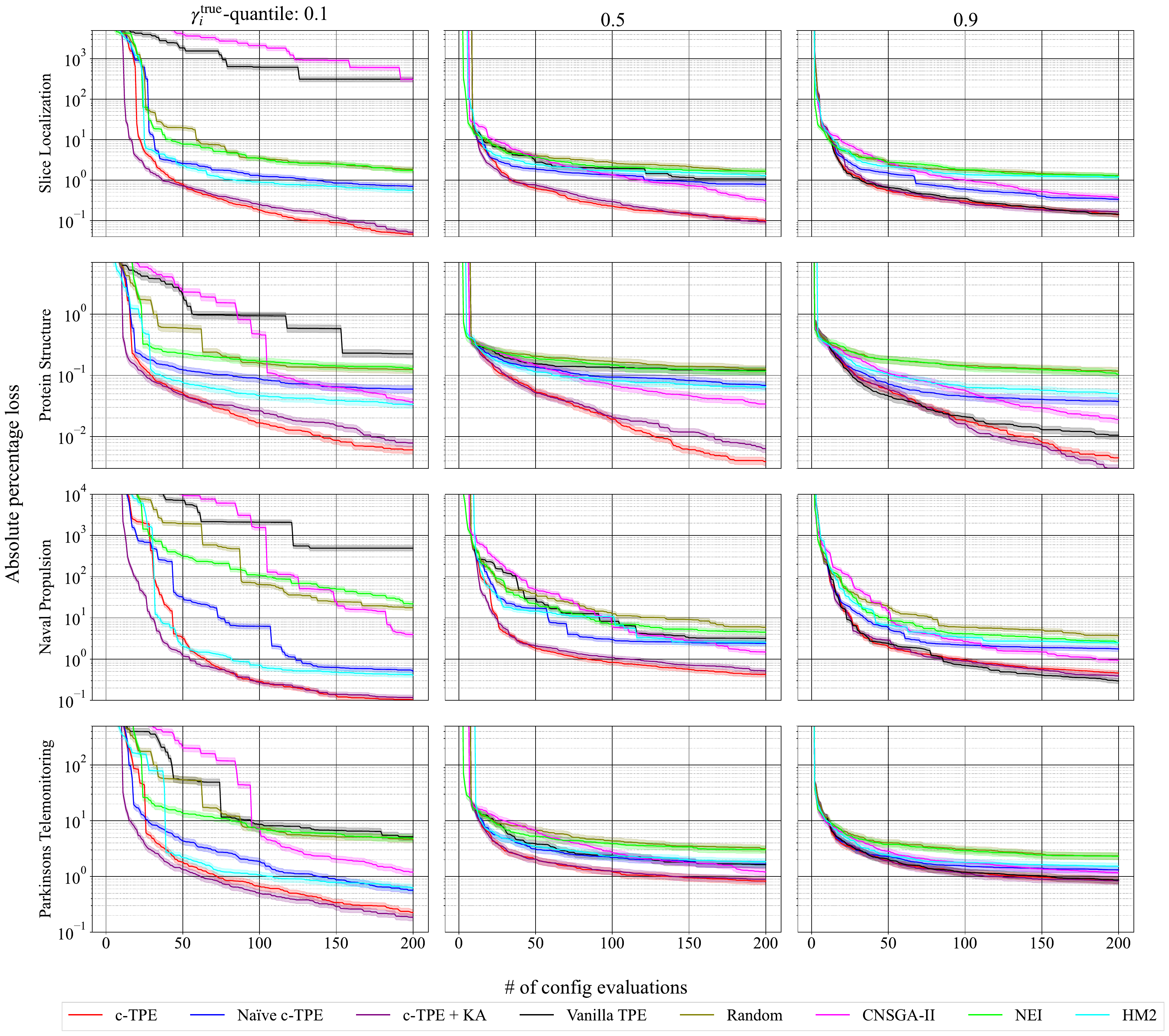}
  \caption{
    The performance curves on four benchmarks in HPOlib 
    with a constraint of network size.
    We picked $\truegamma = 0.1$ (\textbf{Left}),
    $0.5$ (\textbf{Center}), and
    $0.9$ (\textbf{Right}).
    The horizontal axis shows the number of evaluated configurations in optimizations
    and the vertical axis shows the absolute percentage error in each experiment.
  }
  \label{suppl:fig:10-50-90-eval-vs-loss-hpolib-nparams}
  \vspace{-4mm}
\end{figure*}

\begin{figure*}[p]
  \centering
  \includegraphics[width=0.7\textwidth]{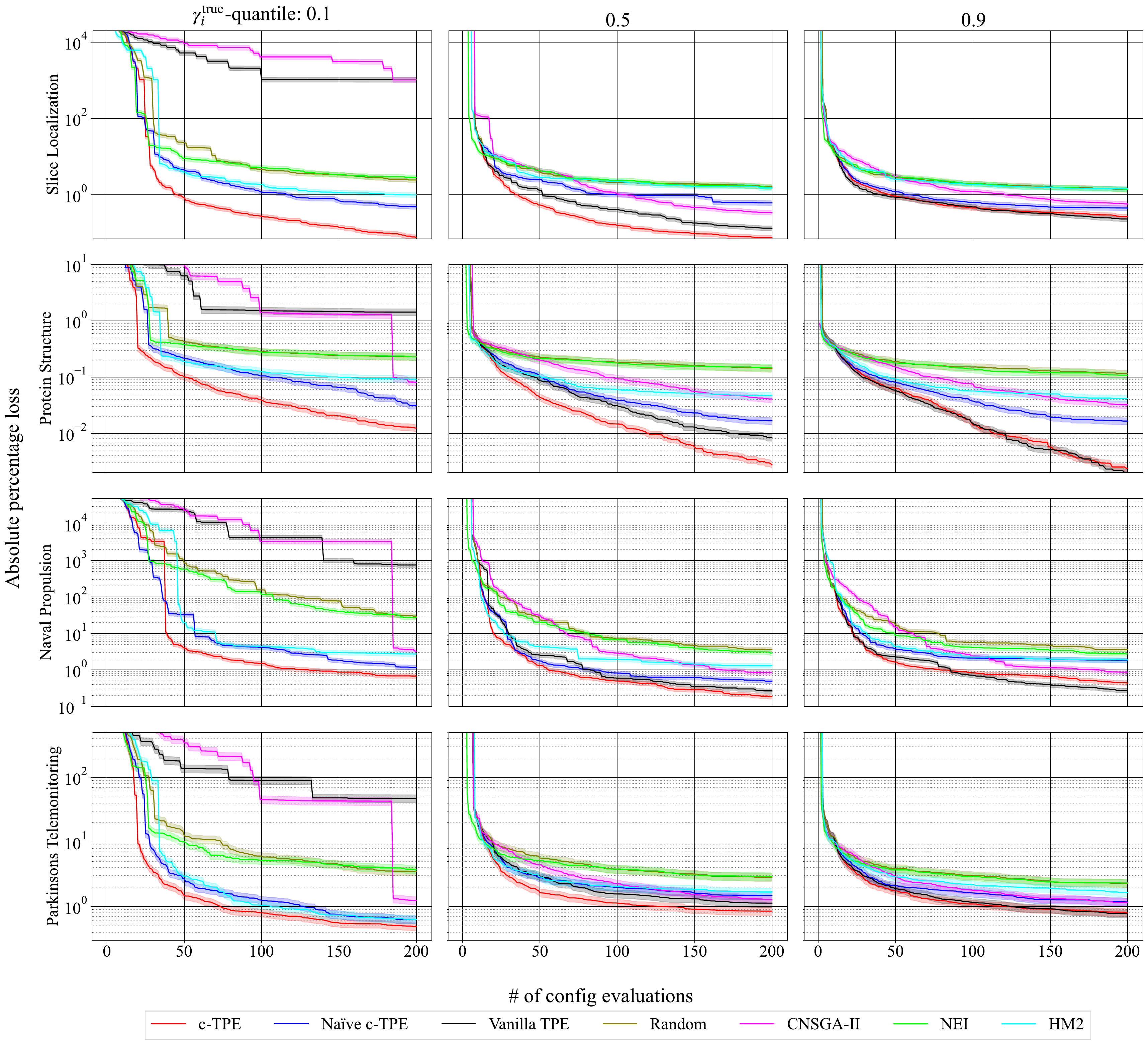}
  \caption{
    The performance curves on four benchmarks in HPOlib 
    with a constraint of runtime.
    We picked $\truegamma = 0.1$ (\textbf{Left}),
    $0.5$ (\textbf{Center}), and
    $0.9$ (\textbf{Right}).
    The horizontal axis shows the number of evaluated configurations in optimizations
    and the vertical axis shows the absolute percentage error in each experiment.
  }
  \label{suppl:fig:10-50-90-eval-vs-loss-hpolib-runtime}
\end{figure*}

\begin{figure*}[p]
  \centering
  \includegraphics[width=0.7\textwidth]{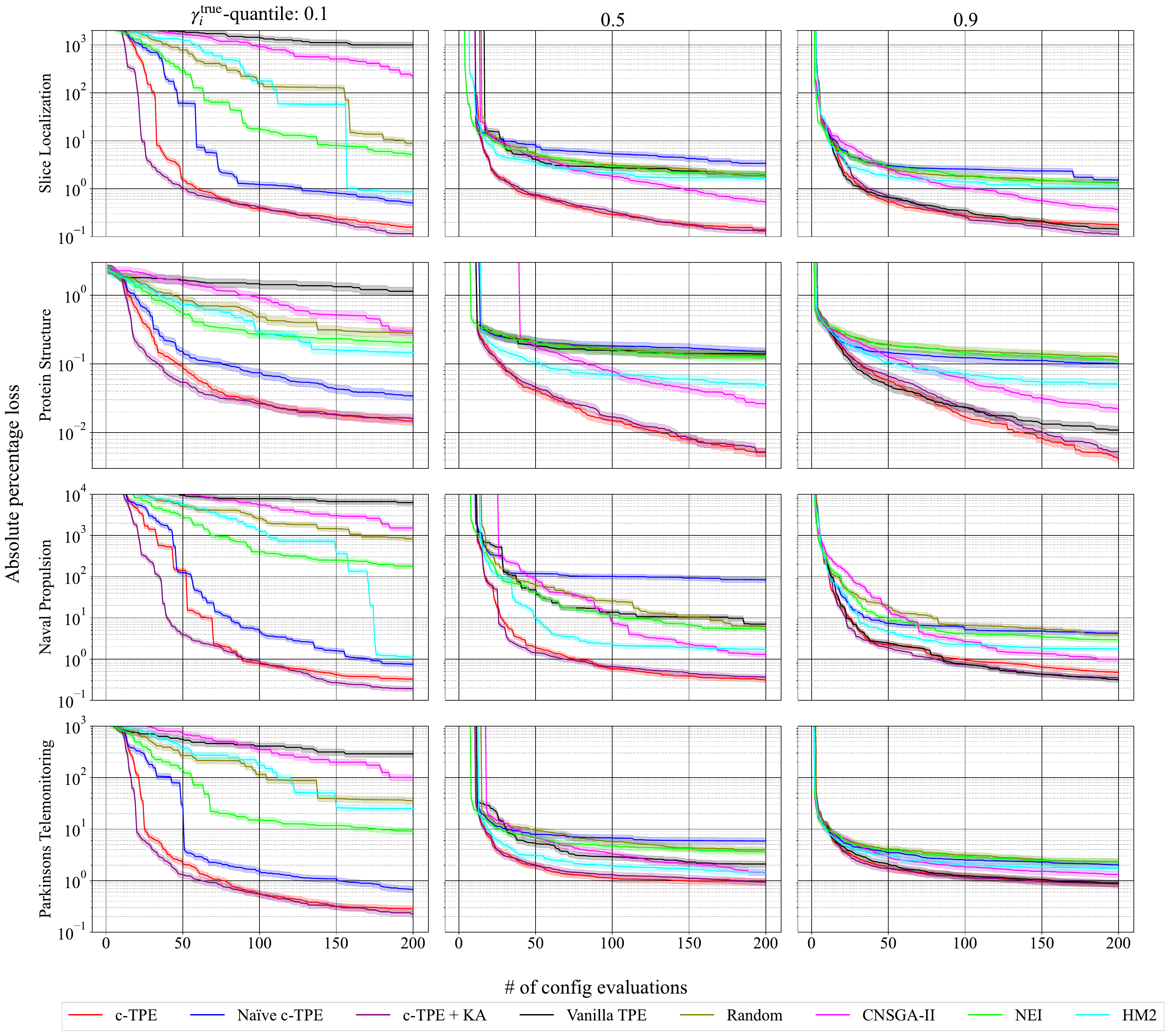}
  \caption{
    The performance curves on four benchmarks in HPOlib 
    with constraints of runtime and network size.
    We picked $\truegamma = 0.1$ (\textbf{Left}),
    $0.5$ (\textbf{Center}), and
    $0.9$ (\textbf{Right}).
    The horizontal axis shows the number of evaluated configurations in optimizations
    and the vertical axis shows the absolute percentage error in each experiment.
  }
  \label{suppl:fig:10-50-90-eval-vs-loss-hpolib-nparams-runtime}
\end{figure*}

Figures~\ref{suppl:fig:10-50-90-eval-vs-loss-hpolib-nparams}--\ref{suppl:fig:10-50-90-eval-vs-loss-hpolib-nparams-runtime}
show the time evolution of absolute percentage loss of each optimization
method on HPOlib with
the $\truegamma$-quantile of 0.1, 0.5, and 0.9.

For the tight constraint settings (\textbf{Left columns}), c-TPE
outperformed other methods and KA accelerated c-TPE in the early stage.
For the loose constraint settings (\textbf{Center, right columns}), CNSGA-II improved
its performance in the early stage of optimizations
although c-TPE still exhibited quicker convergence.
On the other hand, the performance of NEI and HM2 was degraded in the settings of $\truegamma = 0.9$ (\textbf{Right columns})
although such degradation did not happen to c-TPE due to Corollary~\ref{thm:ctpe-converges-to-single-objective}.
In the same vein, KA did not disrupt the performance of c-TPE.

For multiple-constraint settings shown in Figure~\ref{suppl:fig:10-50-90-eval-vs-loss-hpolib-nparams-runtime},
while both CNSGA-II and HM2 showed slower convergence compared to single constraint settings,
c-TPE also showed quicker convergence in the settings.

\subsection{Results on NAS-Bench-101}
\begin{figure*}[p]
  \centering
  \includegraphics[width=0.8\textwidth]{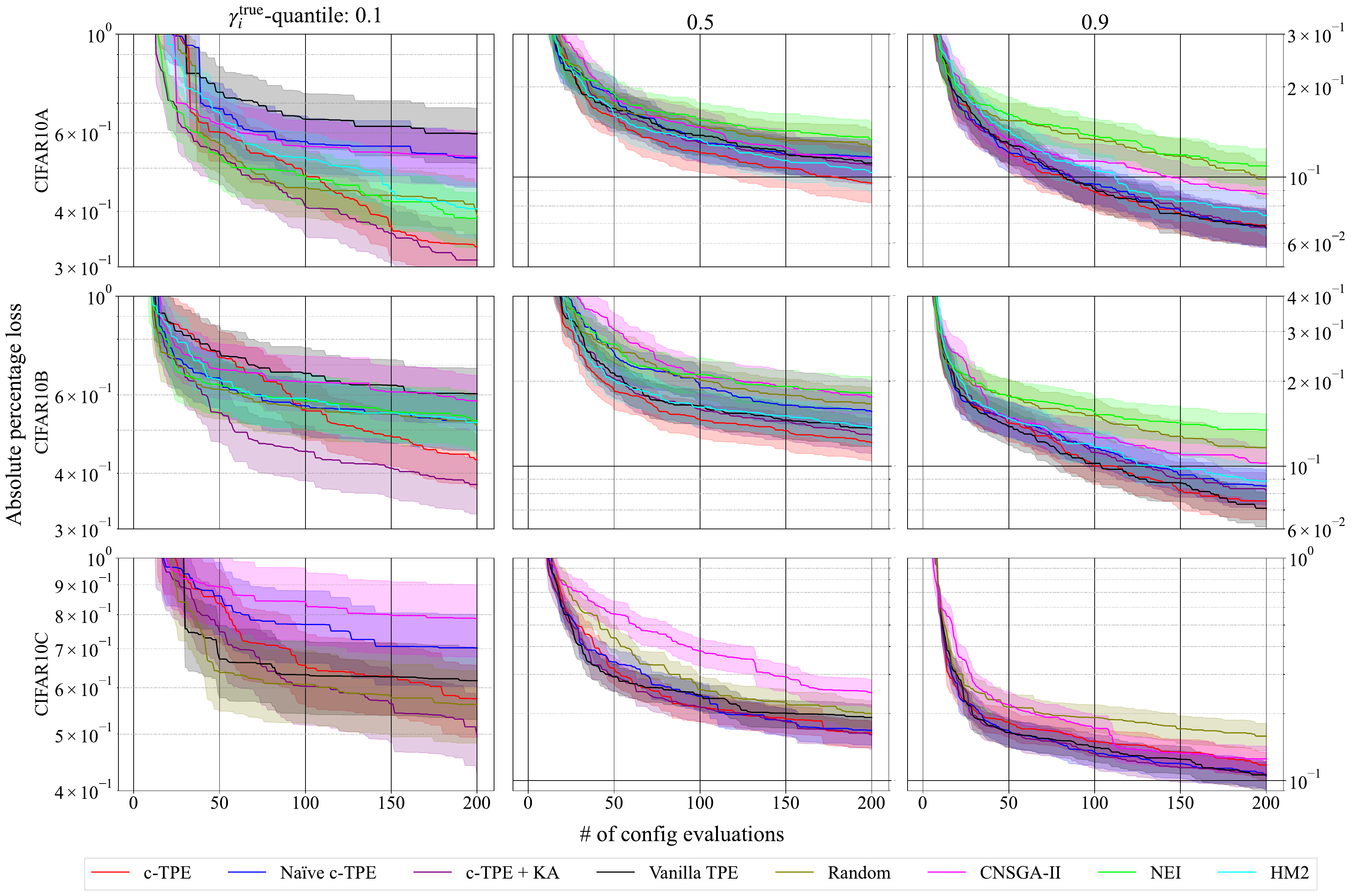}
  \caption{
    The performance curves on three benchmarks in NAS-Bench-101 
    with a constraint of network size.
    We picked $\truegamma = 0.1$ (\textbf{Left}),
    $0.5$ (\textbf{Center}), and
    $0.9$ (\textbf{Right}).
    The horizontal axis shows the number of evaluated configurations in optimizations
    and the vertical axis shows the absolute percentage error in each experiment.
    The scale of the results in 
    $\truegamma = 0.1$ is different from others, we separately scaled for the readability.
  }
  \label{suppl:fig:10-50-90-eval-vs-loss-nb101-nparams}
\end{figure*}

\begin{figure*}[p]
  \centering
  \includegraphics[width=0.8\textwidth]{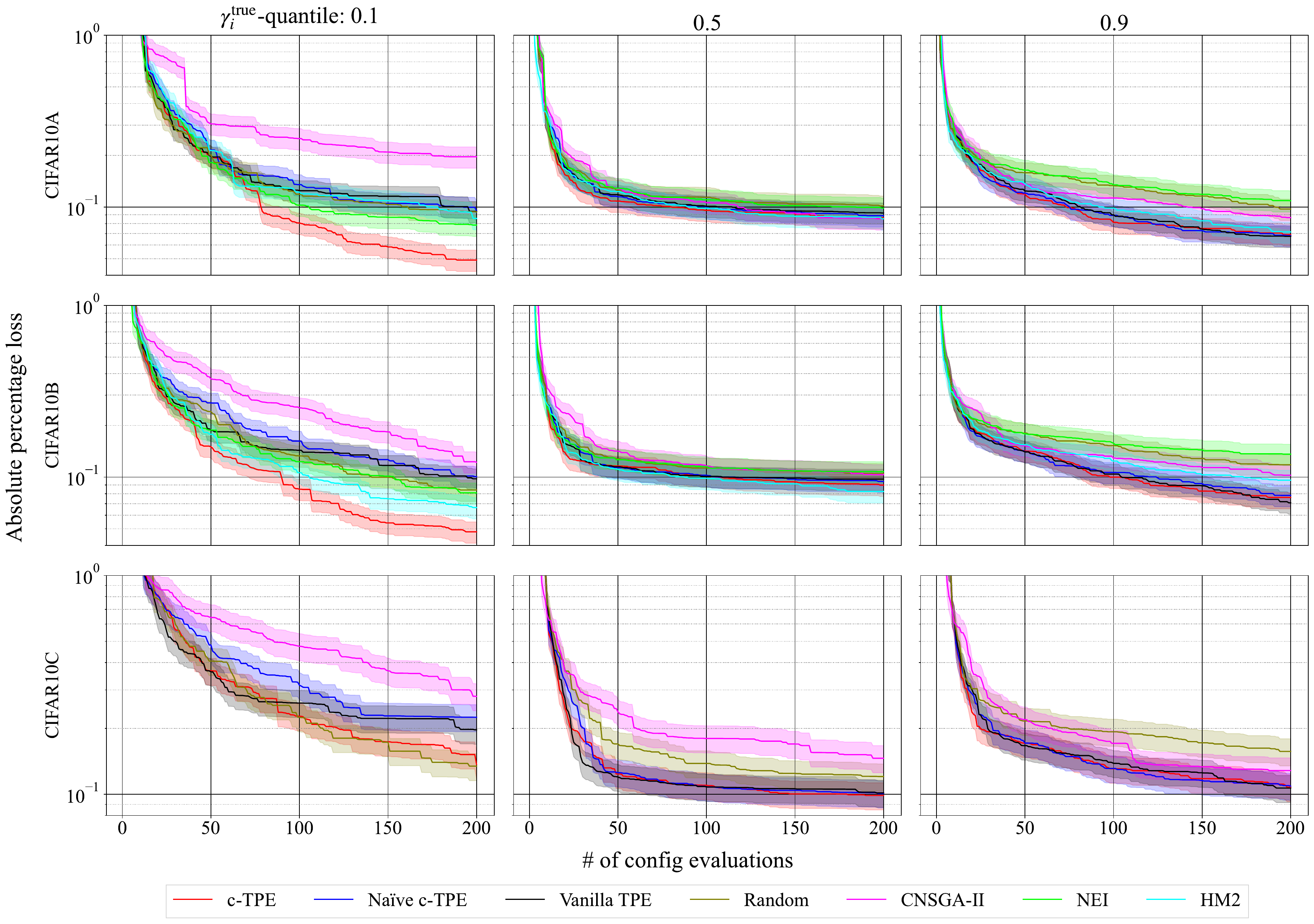}
  \caption{
    The performance curves on three benchmarks in NAS-Bench-101 
    with a constraint of runtime.
    We picked $\truegamma = 0.1$ (\textbf{Left}),
    $0.5$ (\textbf{Center}), and
    $0.9$ (\textbf{Right}).
    The horizontal axis shows the number of evaluated configurations in optimizations
    and the vertical axis shows the absolute percentage error in each experiment.
  }
  \label{suppl:fig:10-50-90-eval-vs-loss-nb101-runtime}
\end{figure*}

\begin{figure*}[p]
  \centering
  \includegraphics[width=0.8\textwidth]{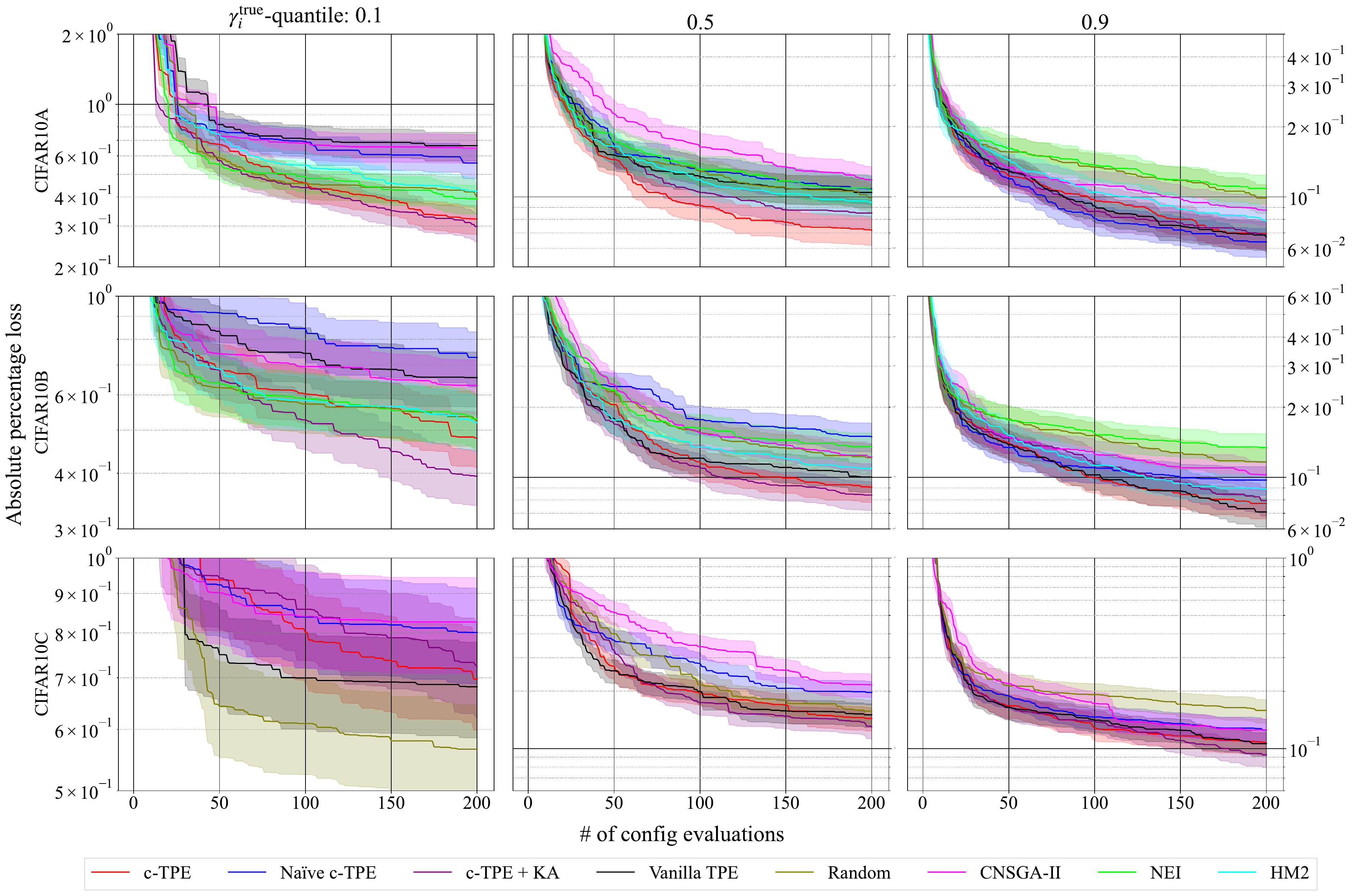}
  \caption{
    The performance curves on three benchmarks in NAS-Bench-101 
    with constraints of runtime and network size.
    We picked $\truegamma = 0.1$ (\textbf{Left}),
    $0.5$ (\textbf{Center}), and
    $0.9$ (\textbf{Right}).
    The horizontal axis shows the number of evaluated configurations in optimizations
    and the vertical axis shows the absolute percentage error in each experiment.
    The scale of the results in 
    $\truegamma = 0.1$ is different from others, we separately scaled for the readability.
  }
  \label{suppl:fig:10-50-90-eval-vs-loss-nb101-nparams-runtime}
\end{figure*}

Figures~\ref{suppl:fig:10-50-90-eval-vs-loss-nb101-nparams}--\ref{suppl:fig:10-50-90-eval-vs-loss-nb101-nparams-runtime}
show the time evolution of absolute percentage loss of each optimization
method on NAS-Bench-101 with
the $\truegamma$-quantile of 0.1, 0.5, and 0.9.
Note that since we could not run NEI and HM2 on CIFAR10C in our environment,
the results for CIFAR10C do not have the performance curves of NEI and HM2.

The results on NAS-Bench-101 look different from those on HPOlib and NAS-Bench-201.
For example, random search outperformed other methods on the tight constraint settings
of CIFAR10C (\textbf{Bottom left}).
Since the combination of high-dimensional search space and tight constraints made the information collection harder,
each method could not guide itself although c-TPE still outperformed other methods on average.
If we add more strict constraints such that c-TPE will pick configurations from feasible domains,
we could potentially achieve better results;
however, as it would lead to poor performance as the number of evaluations increases, this will be a trade-off.
Additionally, KA still helped yield better configurations quickly except CIFAR10C with runtime and network size constraints.
In the loose constraint settings (\textbf{Right column}), since the vanilla TPE exhibited strong performance,
c-TPE also improved its performance in the loose constraint settings due to the effect of Corollary~\ref{thm:ctpe-converges-to-single-objective}.

\subsection{Results on NAS-Bench-201}
\begin{figure*}[p]
  \centering
  \includegraphics[width=0.8\textwidth]{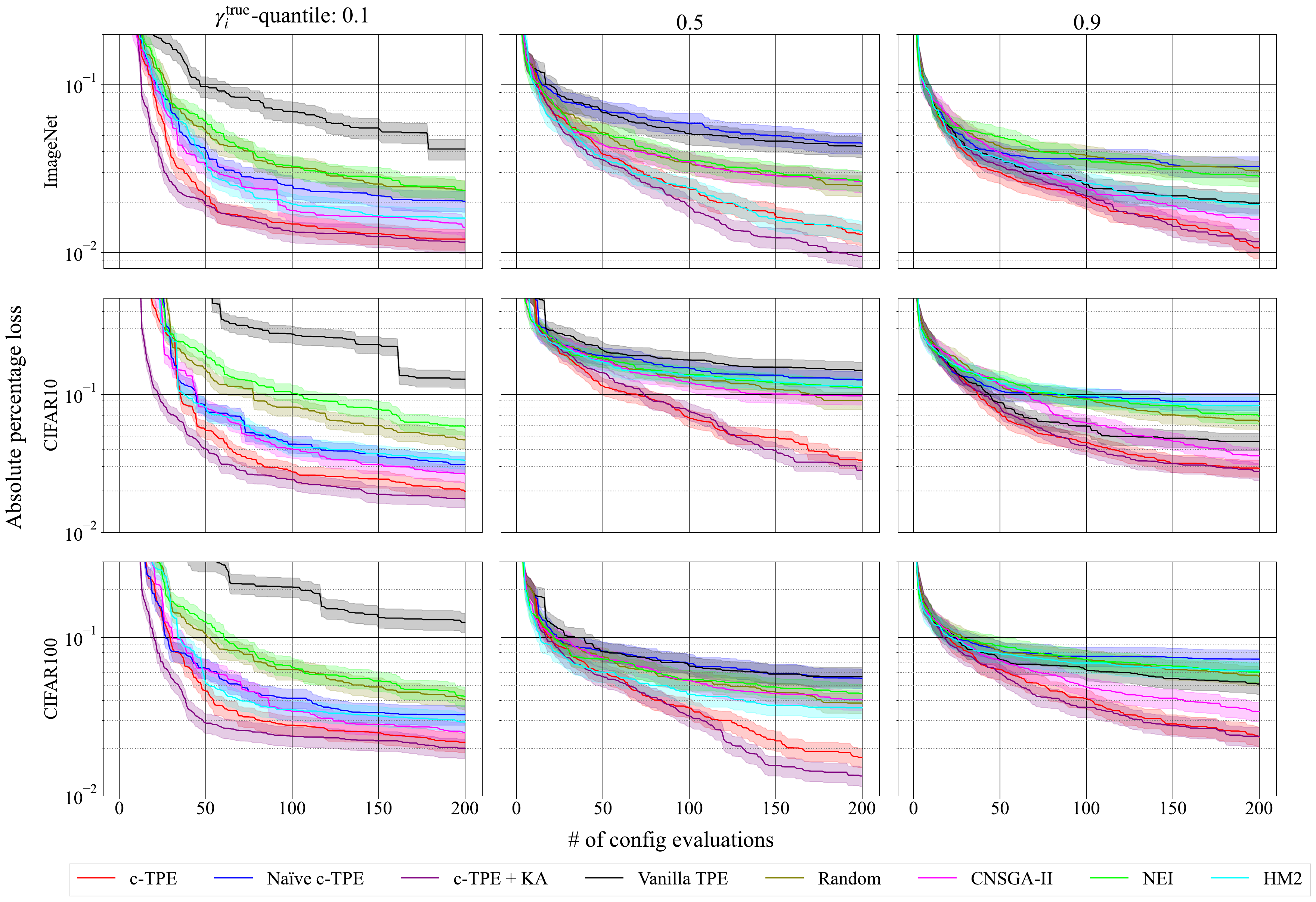}
  \caption{
    The performance curves on three benchmarks in NAS-Bench-201 
    with a constraint of network size.
    We picked $\truegamma = 0.1$ (\textbf{Left}),
    $0.5$ (\textbf{Center}), and
    $0.9$ (\textbf{Right}).
    The horizontal axis shows the number of evaluated configurations in optimizations
    and the vertical axis shows the absolute percentage error in each experiment.
  }
  \label{suppl:fig:10-50-90-eval-vs-loss-nb201-nparams}
\end{figure*}

\begin{figure*}[p]
  \centering
  \includegraphics[width=0.8\textwidth]{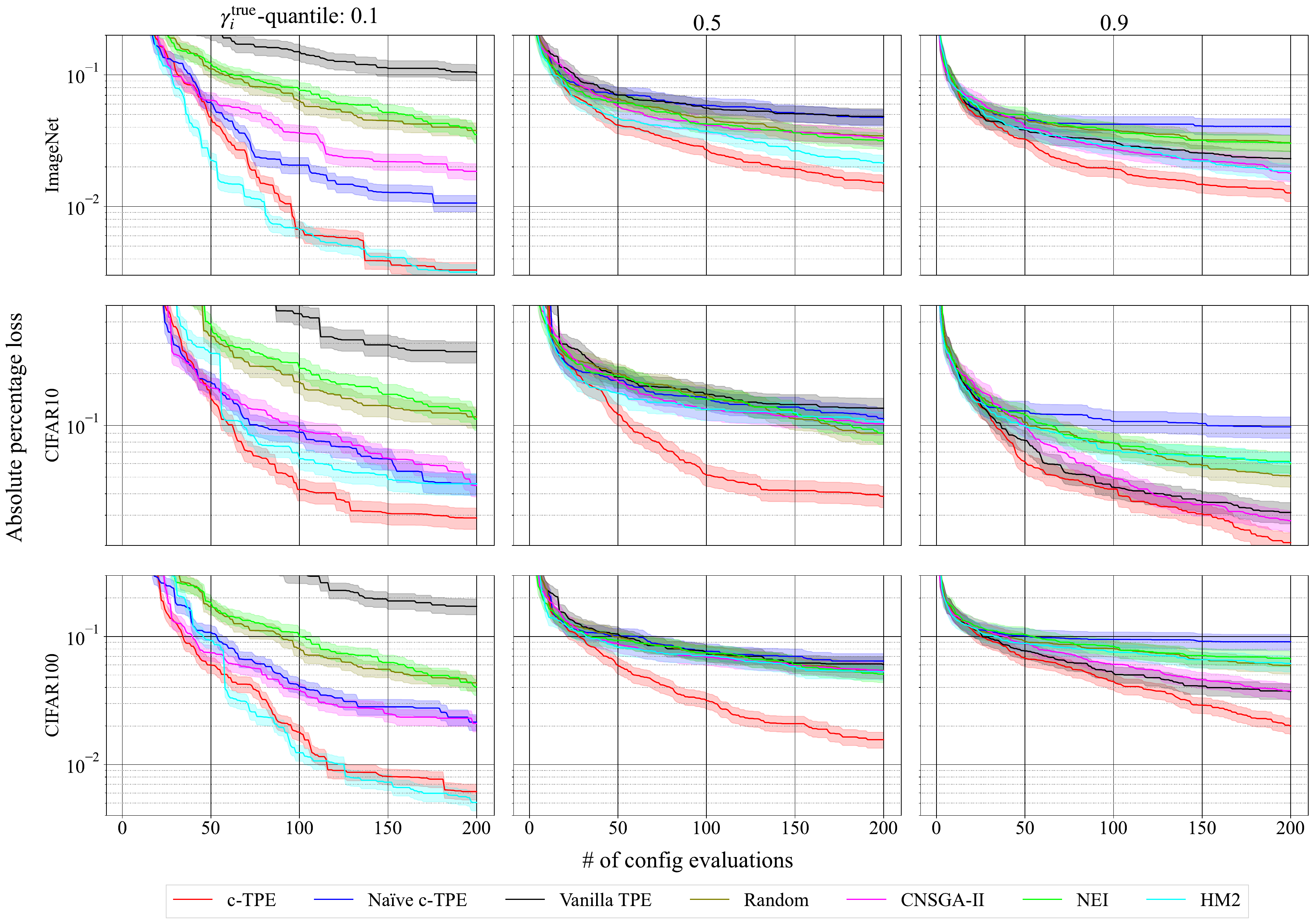}
  \caption{
    The performance curves on three benchmarks in NAS-Bench-201 
    with a constraint of runtime.
    We picked $\truegamma = 0.1$ (\textbf{Left}),
    $0.5$ (\textbf{Center}), and
    $0.9$ (\textbf{Right}).
    The horizontal axis shows the number of evaluated configurations in optimizations
    and the vertical axis shows the absolute percentage error in each experiment.
  }
  \label{suppl:fig:10-50-90-eval-vs-loss-nb201-runtime}
\end{figure*}

\begin{figure*}[p]
  \centering
  \includegraphics[width=0.8\textwidth]{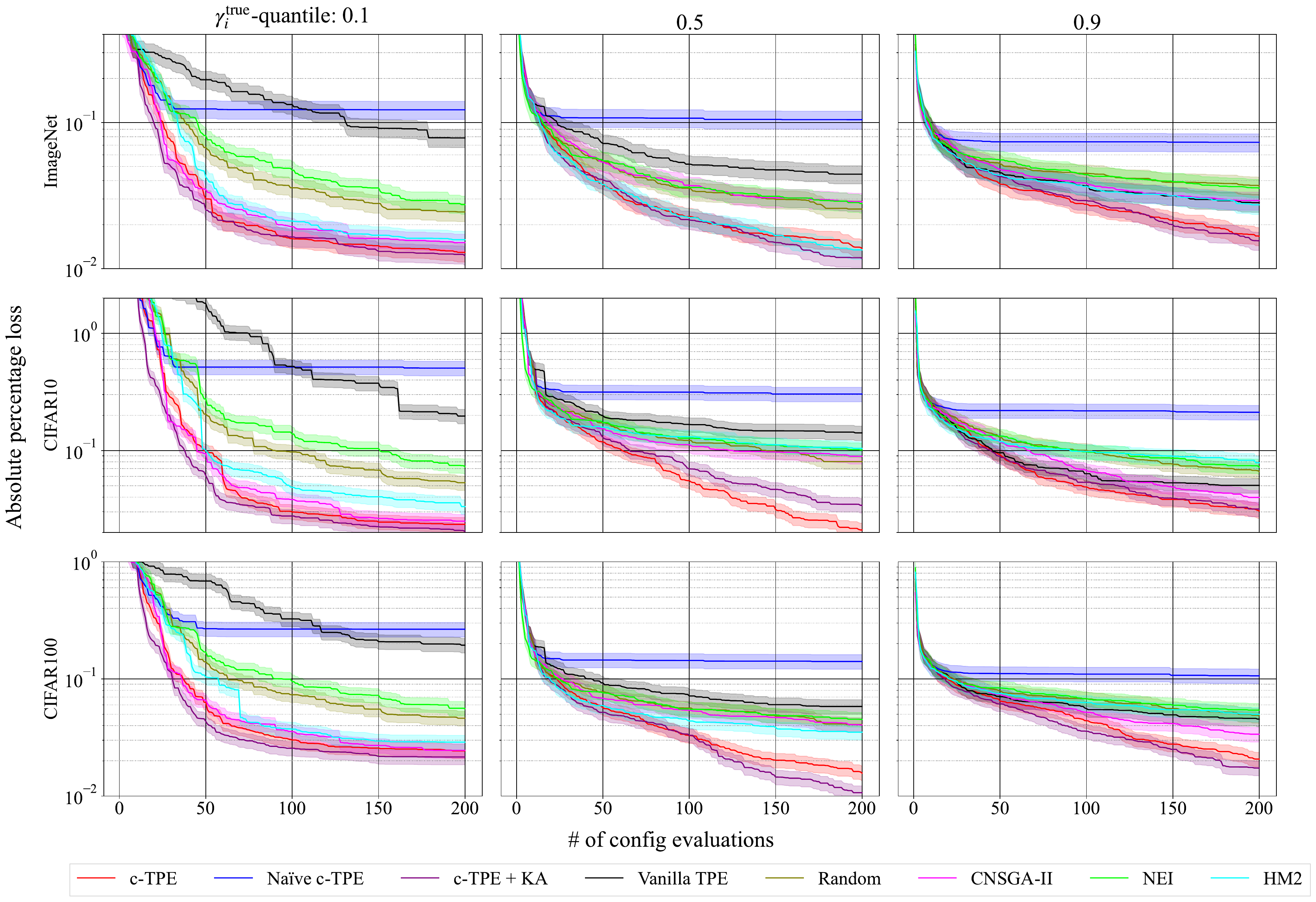}
  \caption{
    The performance curves on three benchmarks in NAS-Bench-201 
    with constraints of runtime and network size.
    We picked $\truegamma = 0.1$ (\textbf{Left}),
    $0.5$ (\textbf{Center}), and
    $0.9$ (\textbf{Right}).
    The horizontal axis shows the number of evaluated configurations in optimizations
    and the vertical axis shows the absolute percentage error in each experiment.
  }
  \label{suppl:fig:10-50-90-eval-vs-loss-nb201-nparams-runtime}
\end{figure*}

Figures~\ref{suppl:fig:10-50-90-eval-vs-loss-nb201-nparams}--\ref{suppl:fig:10-50-90-eval-vs-loss-nb201-nparams-runtime}
show the time evolution of absolute percentage loss of each optimization
method on NAS-Bench-201 with
the $\truegamma$-quantile of 0.1, 0.5, and 0.9.

According to the figures, the discrepancy between c-TPE and the vanilla TPE
is larger than HPOlib and NAS-Bench-101 settings.
This was because of small overlaps discussed in Section~\ref{main:issue02:fig:small-overlap},
and thus the tight constraint settings on NAS-Bench-201 (\textbf{Left columns}) are harder than the other benchmarks.
However, c-TPE and HM2 showed strong performance on the tight constraint settings.
Additionally, c-TPE maintained the performance even over the loose constraint settings (\textbf{Center, right columns}) while CNSGA-II and HM2 did not.
This robustness is also from the property mentioned in Theorem~\ref{main:vanished-constraints:thm:tight-constraint-have-more-priority}.

%% file: appendices/experiments/additional-results-for-rank-over-time.tex
\section{Additional Results for Section~\ref{section:main:experiments:rank-over-time}}
\label{section:suppl:rank-over-time}
\begin{figure*}[p]
  \centering
  \includegraphics[width=0.8\textwidth]{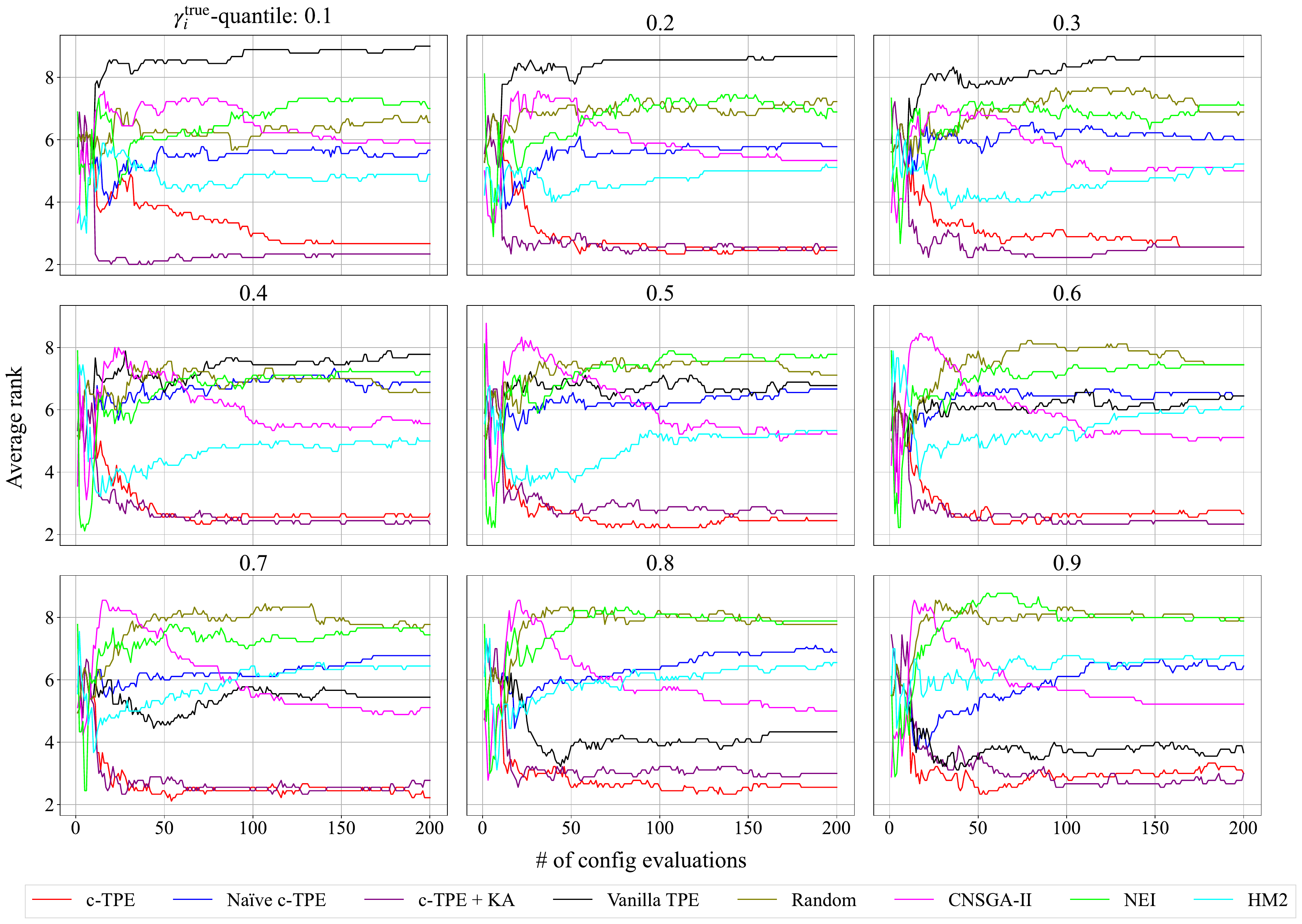}
  \caption{
    The average rank of each method over the number of evaluations.
    We evaluated each method on nine benchmarks with the network size constraint
    and each optimization was repeated over 50 random seeds.
    Each figure presents the results for $\truegamma$ of 0.1 to 0.9, respectively. 
  }
  \label{suppl:fig:eval-vs-avgrank-nparams}
\end{figure*}

\begin{figure*}[p]
  \centering
  \includegraphics[width=0.8\textwidth]{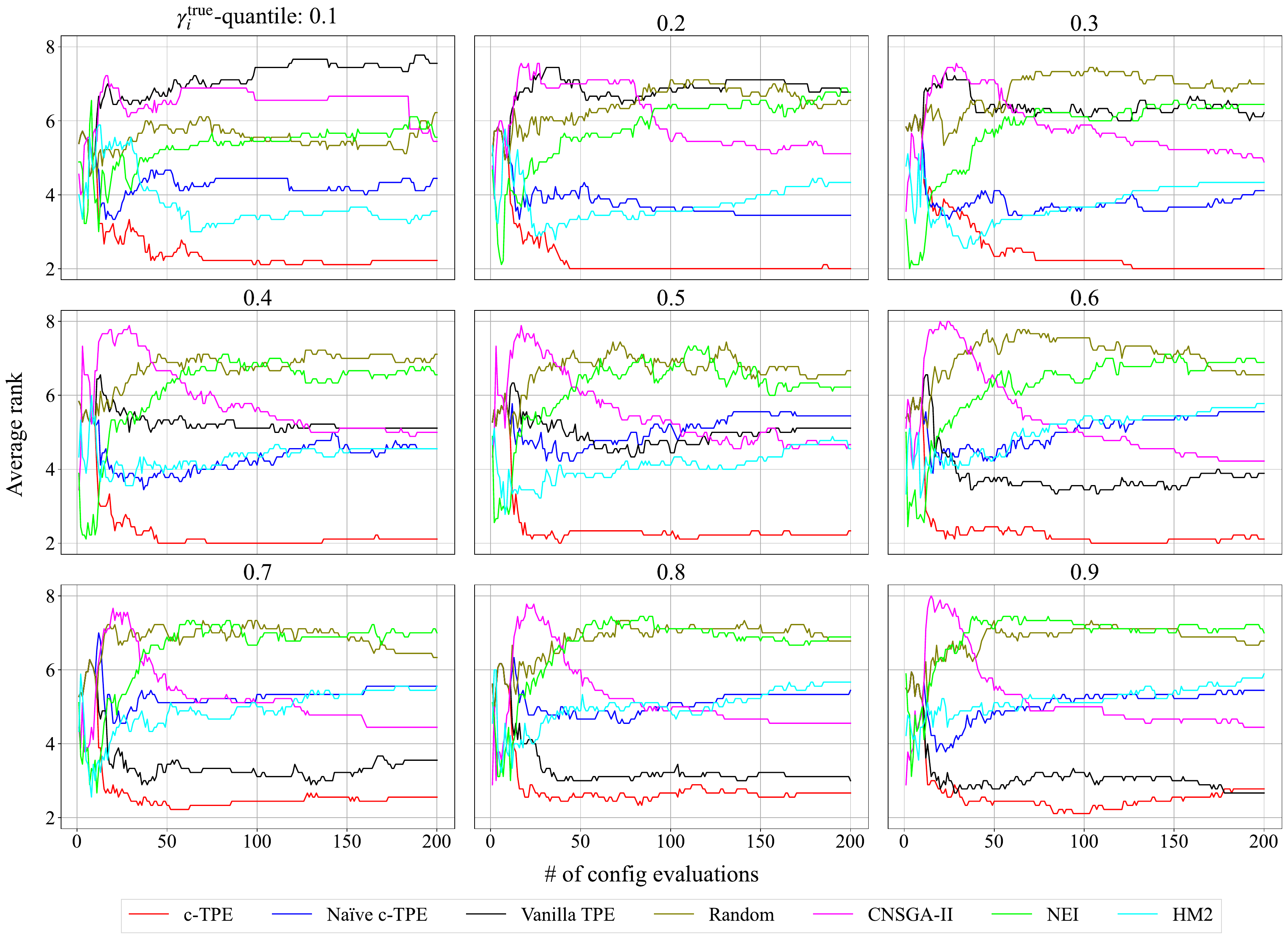}
  \caption{
    The average rank of each method over the number of evaluations.
    We evaluated each method on nine benchmarks with the runtime constraint
    and each optimization was repeated over 50 random seeds.
  }
  \label{suppl:fig:eval-vs-avgrank-runtime}
\end{figure*}

\begin{figure*}[p]
  \centering
  \includegraphics[width=0.8\textwidth]{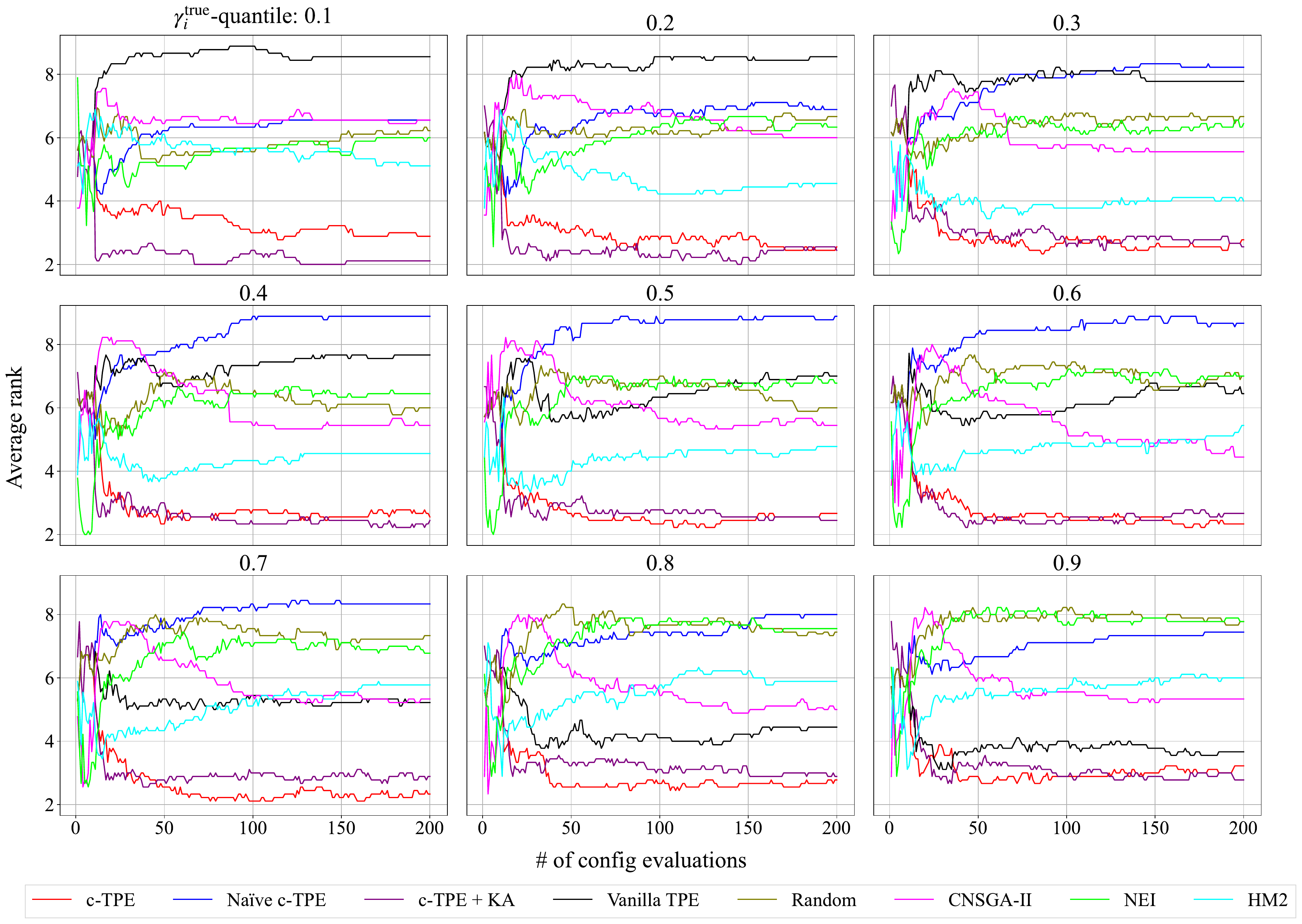}
  \caption{
    The average rank of each method over the number of evaluations.
    We evaluated each method on nine benchmarks with the runtime and the network size constraints
    and each optimization was repeated over 50 random seeds.
  }
  \label{suppl:fig:eval-vs-avgrank-nparams-runtime}
\end{figure*}

Figures~\ref{suppl:fig:eval-vs-avgrank-nparams}--\ref{suppl:fig:eval-vs-avgrank-nparams-runtime}
show the average rank of each method over the number of evaluations.
Each figure shows the performance of different constraint settings with 0.1 to 0.9 of $\truegamma$.

As the constraint becomes tighter, c-TPE converged quicker in the early stage of the optimizations in all the settings due to KA.
On the other hand, KA did not accelerate the optimizations as constraints become looser.
This is because it is easy to identify feasible domains in loose constraint settings even by random samplings.
However, since KA did not degrade the performance of c-TPE, it is recommended to add KA as much as possible.

Furthermore, it is worth noting that although the performance of HM2 and NEI outperformed the vanilla TPE in tight constraint settings,
their performance was degraded as constraints become looser and they did not exhibit better performance than the vanilla TPE with $\truegamma = 0.9$.
On the other hand, c-TPE exhibited better performance than the vanilla TPE even in the settings of $\truegamma = 0.9$
because it adapted the optimization based on the estimated $\hat{\gamma}_i$.

